\documentclass[lettersize,journal]{IEEEtran}
\usepackage{amsmath,amsfonts}
\usepackage{algorithmic}
\usepackage{algorithm}
\usepackage{array}
\usepackage[caption=false,font=normalsize,labelfont=sf,textfont=sf]{subfig}
\usepackage{textcomp}
\usepackage{stfloats}
\usepackage{url}
\usepackage{verbatim}
\usepackage{graphicx}
\usepackage{cite}

\usepackage{booktabs}
\usepackage{multirow}

\usepackage{makecell} % 用于单元格内换行和格式

\usepackage{hyperref}
\usepackage[hyphenbreaks]{breakurl}
\usepackage{cleveref}
\Crefname{figure}{Fig.}{Figs.}

\DeclareRobustCommand*{\IEEEauthorrefmark}[1]{%
    \raisebox{0pt}[0pt][0pt]{\textsuperscript{\footnotesize\ensuremath{#1}}}}

\begin{document}

\title{F{\huge LYING}T{\huge RUST}: A Benchmark for Quadrotor Navigation Across Scenarios and Vehicles}

\author{Gang Li\IEEEauthorrefmark{1}, Chunlei Zhai\IEEEauthorrefmark{1}, Teng Wang\IEEEauthorrefmark{1}, Shaun Li\IEEEauthorrefmark{1}, Shangsong Jiang\IEEEauthorrefmark{2}, and Xiangwei Zhu\IEEEauthorrefmark{1*}~\IEEEmembership{Member,~IEEE}
        % <-this % stops a space
\thanks{\IEEEauthorrefmark{1}School of Electronics and Communication Enginnering, Sun Yet-sen University, Shenzhen, China}% <-this % stops a space
\thanks{\IEEEauthorrefmark{2}Tianwei Xunda (Hunan) Technology Co., Ltd., Hunan, China}% <-this % stops a space
\thanks{\IEEEauthorrefmark{*}Corresponding author: Xiangwei Zhu}% <-this % stops a space
\thanks{Manuscript received XX, 2025; revised XX, 2025.}
\thanks{The benchmark of FLYINGTRUST is available at \url{https://github.com/GangLi-SYSU/FLYINGTRUST}.}}

\markboth{Journal of \LaTeX\ Class Files,~Vol.~14, No.~8, August~2021}%
{Shell \MakeLowercase{\textit{et al.}}: A Sample Article Using IEEEtran.cls for IEEE Journals}

\IEEEpubid{0000--0000/00\$00.00~\copyright~2021 IEEE}

\maketitle

\begin{abstract}
Visual navigation algorithms for quadrotors often exhibit a large variation in performance when transferred across different vehicle platforms and scene geometries, which increases the cost and risk of field deployment. To support systematic early-stage evaluation, we introduce FLYINGTRUST, a high-fidelity, configurable benchmarking framework that measures how platform kinodynamics and scenario structure jointly affect navigation robustness. FLYINGTRUST models vehicle capability with two compact, physically interpretable indicators: maximum thrust-to-weight ratio and axis-wise maximum angular acceleration. The benchmark pairs a diverse scenario library with a heterogeneous set of real and virtual platforms and prescribes a standardized evaluation protocol together with a composite scoring method that balances scenario importance, platform importance and performance stability. We use FLYINGTRUST to compare representative optimization-based and learning-based navigation approaches under identical conditions, performing repeated trials per platform-scenario combination and reporting uncertainty-aware metrics. The results reveal systematic patterns: navigation success depends predictably on platform capability and scene geometry, and different algorithms exhibit distinct preferences and failure modes across the evaluated conditions. These observations highlight the practical necessity of incorporating both platform capability and scenario structure into algorithm design, evaluation, and selection, and they motivate future work on methods that remain robust across diverse platforms and scenarios.
\end{abstract}

\begin{IEEEkeywords}
Benchmark, Quadrotor, Visual navigation, Kinodynamics, Simulation
\end{IEEEkeywords}

\section{Introduction}
\label{Introduction}
\IEEEPARstart{U}{nmanned} Aerial Vehicles (UAVs) are aircraft operated without onboard human pilots, either by remote control or by preprogrammed flight plans~\cite{bharati2018real}. Quadrotors form a particularly prominent class of rotary-wing platforms. By independently modulating the speeds of four motor-propeller units, a quadrotor can generate collective thrust for vertical motion and differential thrust and reaction torques for attitude control. These capabilities enable six degrees of freedom motion combined with fine low-speed control, which drive extensive adoption of quadrotors in precision agriculture, infrastructure inspection, high-resolution mapping, environmental monitoring and disaster response~\cite{amin2016review,ahmed2022recent,kim2018slam,kahn2018self,luo2017distributed,liu2018research,marin2018global,ruiz2015scene,harrison2008high,zhang2017deep}. Central to autonomous operation is visual navigation, a pipeline that integrates camera-based perception, localization and mapping, trajectory planning and continuous control. Over the last decade, many high-performance visual navigation methods have been developed, ranging from classical optimization-based planners to recent learning-based approaches~\cite{fast-planner,zhou2020ego,loquercio2021learning,xu2025navrl}.

\begin{figure}[!t]
\centering
\includegraphics[width=3.4in]{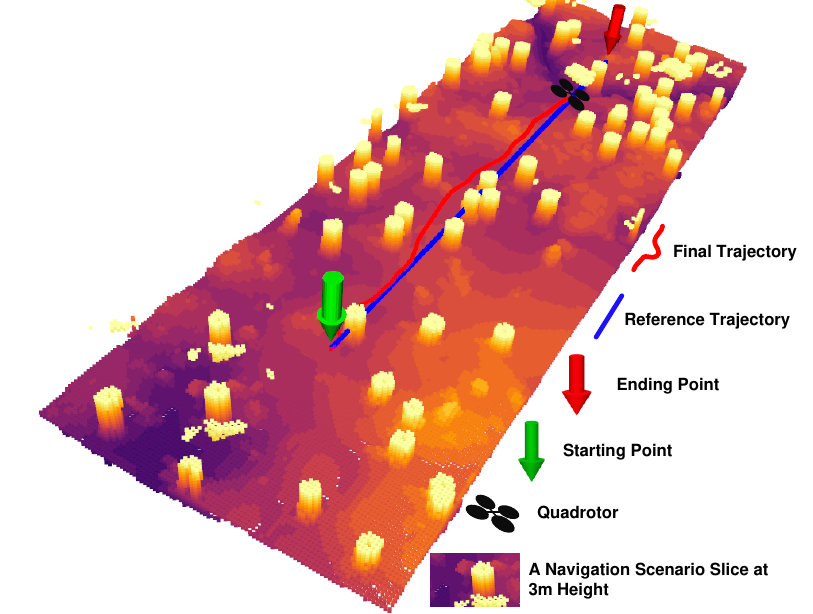}
\caption{{\bf{Representative benchmark scenario.}} The environment and obstacle layout, showing the quadrotor's start (green arrow) and goal (red arrow) positions. The blue line represents the straight-line reference path, and the red curve is an example of a collision-free trajectory executed by a planner.}
\label{navigation_visualization}
\end{figure}
\IEEEpubidadjcol
In practice, however, one same visual navigation algorithm can produce markedly different outcomes when deployed on different quadrotor platforms or in different environments. Many algorithms are developed and tuned under implicit assumptions about actuator effectiveness, sensor latency, control bandwidth and vehicle agility~\cite{foehn2022agilicious}. When those assumptions are not satisfied on a target platform, or when the geometric demands of the environment change, the algorithm’s success rate can drop sharply. Typical failure modes include executing turns too early or too late, becoming trapped in dead ends, or violating kinodynamic constraints and entering unstable flight regimes. These failures substantially increase the time, cost and risk associated with real-world field trials and complicate decisions about which algorithms merit further investment in hardware testing. Despite its practical importance, the community lacks a systematic, reproducible tool for evaluating how scenario structure and platform performance jointly affect navigation robustness during the algorithm development stage.

To address this requirement we present {\bf{FLYINGTRUST}}, a configurable simulation-based benchmark for quadrotor visual navigation. As illustrated in \Cref{navigation_visualization}, FLYINGTRUST is explicitly designed to run in simulation as a proactive, low-risk evaluation stage that guides subsequent real-world testing. The benchmark identifies brittle algorithm-platform-scenario interactions so that researchers and practitioners can prioritize which algorithm-platform pairs deserve costly field trials, and it does so without requiring any new physical flight experiments. This simulation-first design positions the benchmark as a development and prioritization tool, rather than as a replacement for final real-world validation. The benchmark pursues three complementary aims. First, it quantifies navigation success as a function of both environment characteristics and platform capability, thereby making brittle interactions visible early. Second, it exposes common, interpretable failure modes that arise from specific platform limitations, for example insufficient thrust-to-weight ratio or limited angular acceleration about particular axes. Third, it provides a standardized, reproducible evaluation protocol so that different research groups can compare methods fairly and prioritize which algorithm-platform pairs should advance to field testing.

FLYINGTRUST implements the benchmark as an end-to-end simulation pipeline. First, we define a compact, parametric representation of quadrotor performance and construct a fixed, heterogeneous set of platform profiles, including documented real platforms and a set of virtual platforms obtained by interpolation within the documented design space. Next, we assemble a fixed scenario library that contains both common navigation scenes and targeted stress-test environments. The scalable test set is produced by cross-joining the selected platform profiles with the scenario instances to generate all platform-scene combinations to be evaluated. For each platform-scene pair, the framework runs repeated trials under standardized task specifications and flight limits, collects success and diagnostic metrics, and computes uncertainty-aware summary statistics. Finally, results are aggregated via a principled composite score that weights scenario importance, platform importance, and performance stability to produce an intuitive ranking and to highlight brittle algorithm-platform-scene interactions for follow-up field evaluation.

In this paper we adopt a simulation-first benchmarking framework, FLYINGTRUST, to systematically evaluate how environmental geometry and platform kinodynamic capability jointly influence the robustness of visual navigation algorithms. The benchmark pairs 18 curated real-world quadrotor platforms with 18 virtual platforms generated by interpolation in the documented design space, and evaluates navigation across seven representative scenes. This yields 252 platform-scene combinations; each combination is tested across multiple independent trials to report success rates, confidence intervals and stability metrics. The manuscript integrates method description, experimental protocol and analysis: \Cref{Taxonomy} classifies current visual quadrotor navigation approaches, \Cref{FLYINGTRUST} describes the FLYINGTRUST design and implementation, \Cref{Experiments} presents the experimental setup and raw results, provides statistical analysis and practical recommendations for algorithm selection, \Cref{Broader Impact and Limitations} discusses limitations and directions for future extensions. FLYINGTRUST is intended to make brittle algorithm-platform-scene interactions visible in simulation so researchers can prioritize which algorithm-platform pairs merit costly real-world flight tests, thereby reducing development cost and risk.

\section{Landscape of Visual Quadrotor Navigation Methods}
\label{Taxonomy}

\begin{figure*}[!ht]
\centering
\includegraphics[width=6.5in]{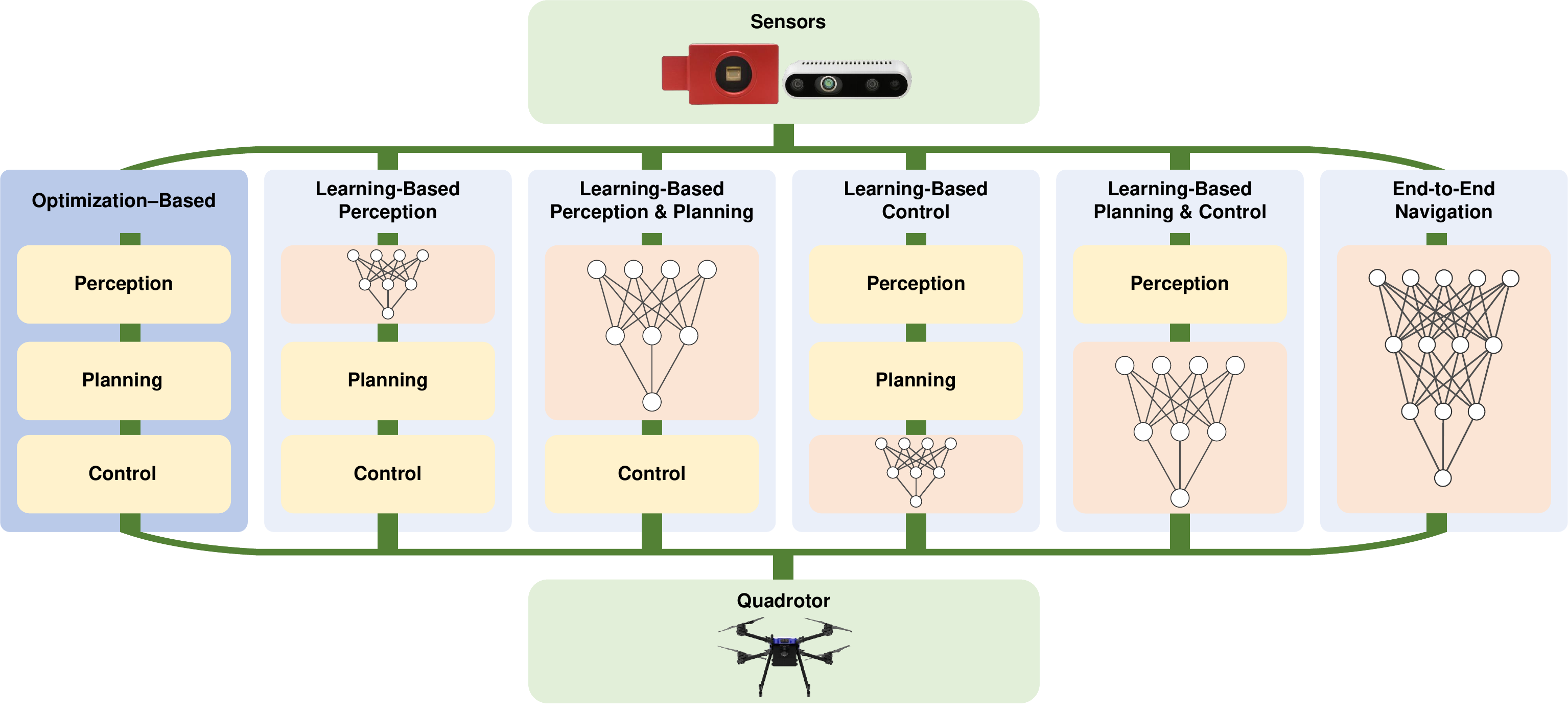}
\caption{{\bf{Taxonomy of quadrotor visual navigation algorithms.}} A flowchart of common navigation pipelines, from sensors to quadrotor. The figure contrasts the modular, optimization-based pipeline (left) with various learning-based approaches, which are categorized by the components replaced with neural networks (e.g., perception, planning, control).}
\label{two categories}
\end{figure*}

Most modern visual navigation systems for quadrotors follow a three-stage paradigm~\cite{hanover2024autonomous}, with some exceptions that adopt fully end-to-end learning. In the dominant approach, perception produces information from onboard sensors and a planner uses that information to generate a collision-free trajectory, after which a controller tracks the planned trajectory on the physical vehicle. This design pattern highlights two interdependent questions that determine real-world performance. First, given diverse and challenging environments, can the planner reliably produce collision-free, kinodynamically-feasible trajectories? Second, once such a trajectory is available, can the controller faithfully execute it on a particular quadrotor platform under its actuator and inertia limits? The benchmark proposed in this paper explicitly targets these two questions by evaluating planners and controllers jointly across a range of scene geometries and platform performance profiles.

As illustrated in \Cref{two categories}, we now summarize the principal methodological families encountered in the literature, describe their characteristic strengths and limitations, and comment on the kinds of environments and vehicle capabilities to which they are best suited.

\subsection{Optimization-based}

Conventional optimization-based pipelines separate perception and mapping, trajectory optimization, and robust control. The typical workflow builds a local environment representation or distance field, parameterizes a trajectory (often with splines), and solves a constrained optimization that balances smoothness, collision avoidance and dynamic feasibility. The main strengths of these methods are interpretability and the ability to enforce kinodynamic constraints explicitly during planning. They therefore suit structured or moderately cluttered environments where reliable local maps and distance information are available, and they benefit platforms that provide sufficient thrust authority and angular acceleration to realize aggressive trajectories. Their primary limitations are computation and mapping cost on resource-constrained platforms, and susceptibility to local minima in gradient-based solvers; topology-aware or multi-start strategies help mitigate the latter. Representative works include real-time replanning using a 3D ring buffer and spline parameterization \cite{usenko2017real}, B-spline trajectory optimization with dynamics-aware initial search \cite{fast-planner}, path-guided optimization that leverages multiple homotopy classes \cite{PGO}, and ESDF-free local replanners that focus computation near a guide path \cite{zhou2020ego}.

\subsection{Learning-based}

Learning-based methods replace one or more traditional modules with neural networks~\cite{lee2021flying,pham2022deep}. Depending on which functions are learned, these methods differ substantially in capabilities and requirements. Below we briefly characterize the main subclasses and note the environments and platform types for which they tend to be most appropriate.

\subsubsection{Learning-Based Perception}

Learned perception modules (for example CNNs) detect task-relevant features or produce compact scene embeddings from RGB, depth or event sensors. When trained appropriately, they can improve robustness to visual degradation and reduce the end-to-end latency of the perception stack. They are most useful in scenarios with challenging lighting or texture conditions where handcrafted detectors fail. Practical constraints include the inference cost on embedded hardware and the need for annotated or high-fidelity simulated training data \cite{foehn2022alphapilot}.

\subsubsection{Learning-Based Perception and Planning}

Replacing mapping and local planning with learned models enables navigation without explicit maps. Image-to-waypoint and short-horizon learned planners provide reactive behavior with low online computation, which is advantageous in high-speed or computation-limited platforms. These methods perform well when training data reflect the test distributions; they are vulnerable to out-of-distribution scene geometry and typically require careful dataset design or domain randomization \cite{kaufmann2018deep,loquercio2021learning}.

\subsubsection{Learning-Based Control}

Reinforcement learning can produce low-level controllers that map observed states to actuator commands. RL controllers can achieve excellent performance in simulation and may outperform hand-tuned controllers on some metrics. Their drawbacks are limited formal stability guarantees and sensitivity to simulator mismatch, which complicate direct transfer to hardware without additional safety measures \cite{koch2019reinforcement,lambert2019low}.

\subsubsection{Learning-Based Planning and Control}

Methods that jointly learn planning and control produce policies that directly map observations to actions or short trajectories. Such approaches can achieve highly reactive, time-efficient behavior in tasks like racing, provided that training covers the necessary scene and dynamics variability. Their generalization and safety in arbitrary environments remain active research challenges \cite{kaufmann2023champion,penicka2022learning,xu2025navrl}.

\subsubsection{End-to-End Navigation}

End-to-end systems map sensor inputs to control outputs. Two operational variants are common:
\begin{itemize}
  \item \textbf{Modular end-to-end:} learned modules replace individual pipeline blocks (perception, planning, control) and are trained jointly.
  \item \textbf{Fully end-to-end:} a single model maps raw sensors to low-level commands.
\end{itemize}

End-to-end designs can reduce latency and exploit large datasets, but they pose challenges for interpretability, enforcement of hard safety constraints, and cross-platform generalization. Hybrid architectures that combine learned components with optimization-based safety or control layers represent a practical compromise \cite{muller2019learning,muller2018teaching,rojas2020deeppilot}.

Although many quadrotor visual navigation algorithms, both optimization-based and learning-based, report strong performance in simulation, most are developed and tested under simplified or idealized conditions. These conditions often assume near-ideal actuator response, negligible sensor latency, and stable kinodynamics, assumptions that rarely hold in practice. As a result, algorithm performance often varies or degrades significantly across different platforms and environments. Moreover, the literature provides limited systematic analysis of how specific kinodynamic characteristics and scenario structures jointly influence robustness.

\begin{figure*}[t]
\centering
\includegraphics[width=7in]{"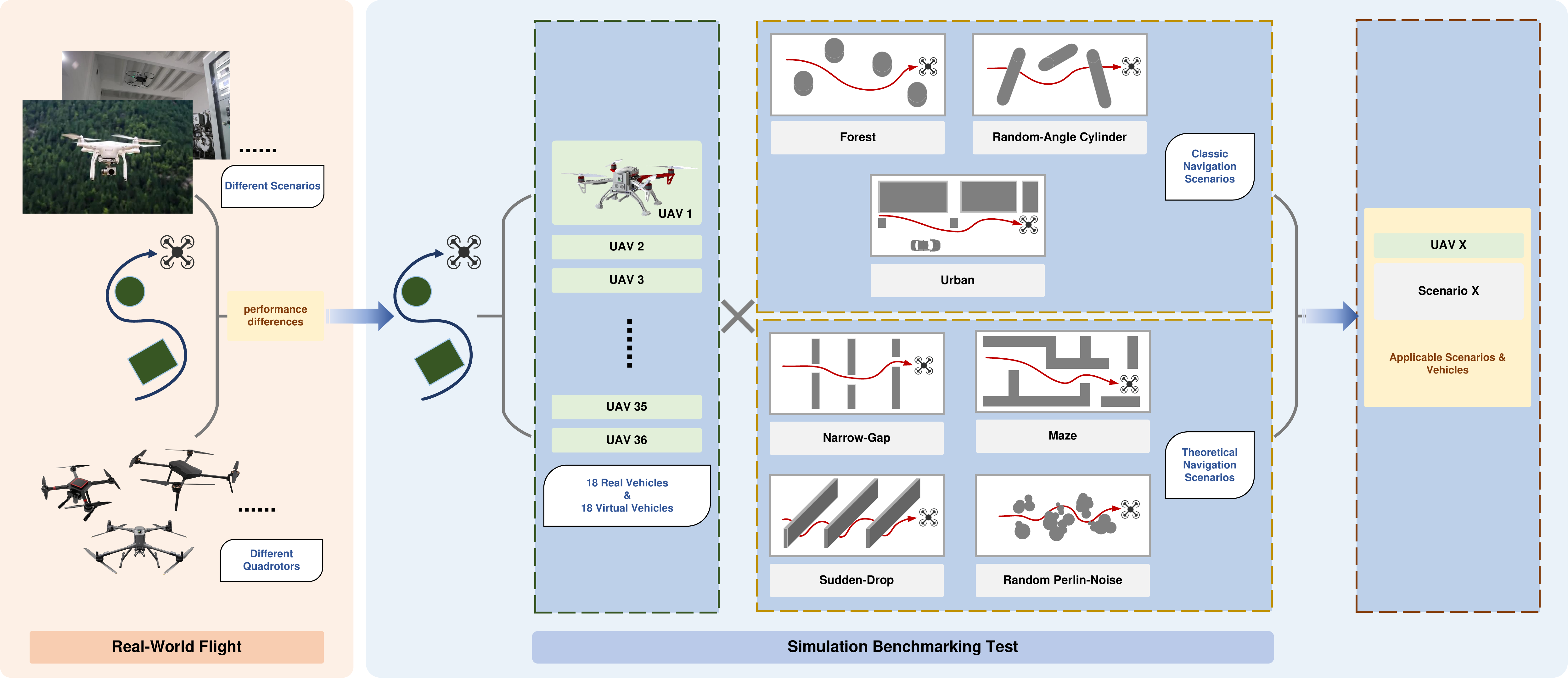"}
\caption{{\bf{Overview of the FLYINGTRUST benchmarking pipeline.}} The benchmark pairs a fixed set of platform profiles with a fixed scenario library to form platform-scenario combinations; each combination is evaluated with multiple trials and summarized via a composite scoring scheme.}
\label{BENCHMARK}
\end{figure*}

To address this gap, we introduce FLYINGTRUST, illustrated in \Cref{BENCHMARK}, a comprehensive and configurable benchmarking framework that enables rigorous and reproducible evaluation of UAV navigation algorithms. FLYINGTRUST explicitly considers both platform performance and scenario geometry, providing a principled way to analyze how these factors interact to shape algorithm robustness. By doing so, it supports fair comparison across methods and offers practical guidance for selecting algorithms suitable for specific UAV platforms and navigation scenarios.

\section{FLYINGTRUST}
\label{FLYINGTRUST}
\subsection{Kinodynamic performance of quadrotors}

\begin{figure*}[!ht]
\centering
\includegraphics[width=5.5in]{"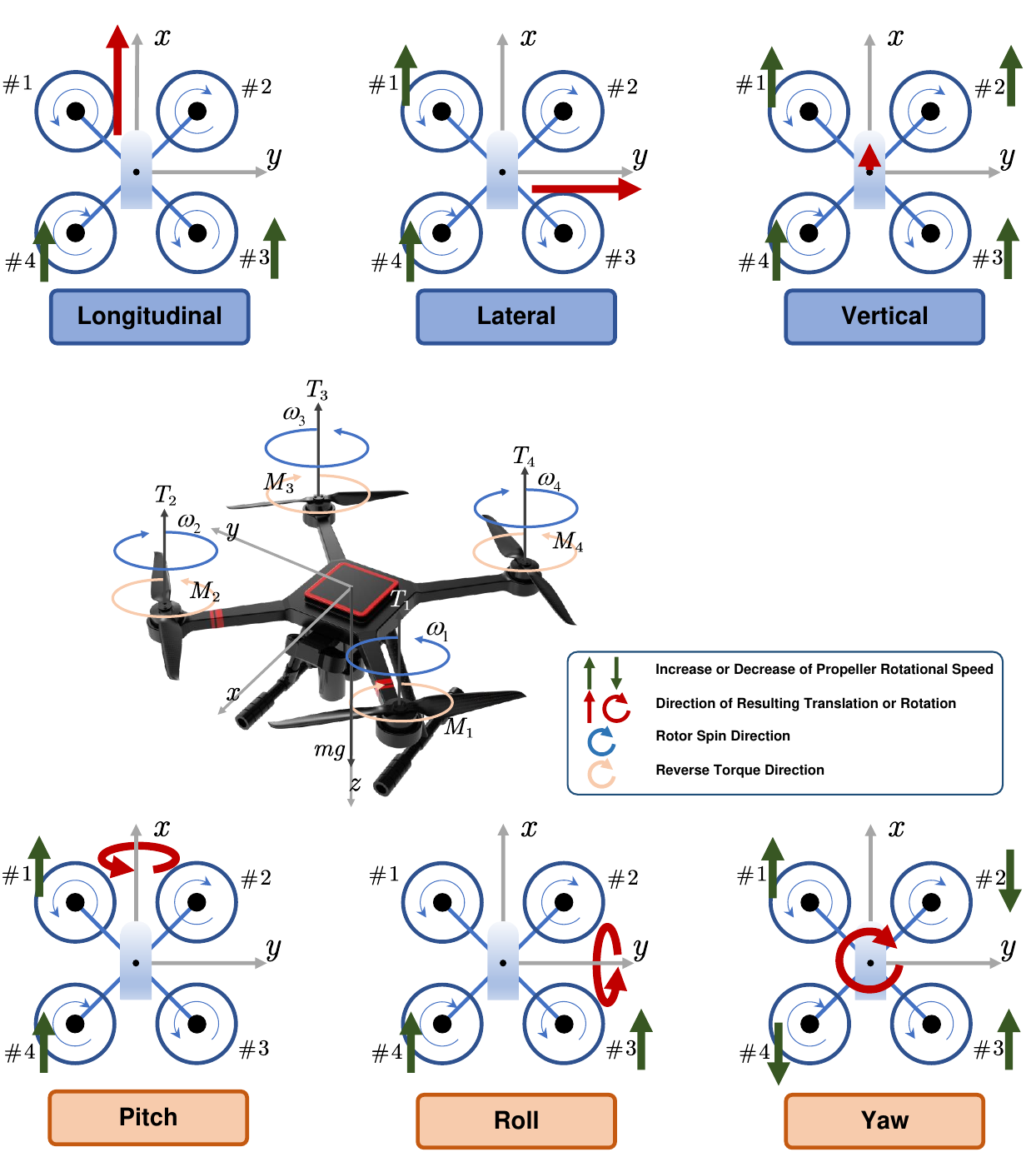"}
\caption{{\bf{Illustration of how rotor speed modulation produces the six degrees of freedom motions of a quadrotor.}} The central diagram illustrates forces (e.g., $T_i$, mg) and torques ($M_i$) on the vehicle. The surrounding schematics show the required changes in rotor speed (green arrows) to achieve the six canonical motions: longitudinal, lateral, vertical, pitch, roll, and yaw (red arrows).}
\label{Six-degree-of-freedom motion of a quadrotor}
\end{figure*}

A quadrotor is a typical six-degree-of-freedom vertical take-off and landing (VTOL) aircraft~\cite{achtelik2009autonomous}. As illustrated in \Cref{Six-degree-of-freedom motion of a quadrotor}, by adjusting the rotational speeds of its four motors, a quadrotor can perform vertical motion along the z axis, forward and backward motion along the x axis, lateral motion along the y axis, as well as rotational motions including pitch (around the y axis), roll (around the x axis), and yaw (around the z axis)~\cite{quan2017introduction}.

In extensive engineering practice we observe that quadrotor flight frequently involves unsteady, time varying maneuvers such as acceleration, deceleration, climbing, descending, orbiting and aggressive acrobatics. These behaviors reflect two related but distinct aspects of flight dynamics:

\begin{itemize}
  \item {\bf{Maneuverability}} denotes the vehicle’s ability to change its speed, altitude or heading over a finite time horizon and can be decomposed into speed, altitude and directional maneuverability.
  \item  {\bf{Agility}} emphasizes transient response, that is, how quickly and precisely the vehicle can switch between motion states; typical facets are roll agility, pitch agility and yaw agility.
\end{itemize}

Although some studies report metrics such as “maximum turn rate” or short-duration aggressive turn performance~\cite{machmudah2022flight}, there is little consistency in how sustained, continuous turn rate versus transient peak turn behavior is defined or used. These metrics tend to be numerous, fragmented, and difficult to compare across platforms and scenarios. From rigid body dynamics we note that linear accelerations and angular accelerations are determined directly by the net forces and torques acting on the vehicle, and these accelerations govern position and attitude evolution. Therefore, we adopt two compact, physically interpretable indicators to characterize kinodynamic performance: The {\bf{maximum thrust to weight ratio (TWR)}}, which primarily quantifies maneuverability, and the {\bf{maximum angular accelerations about the three body axes}}, which capture agility.

{\bf{Maximum Thrust-to-Weight Ratio (TWR):}} The thrust-to-weight ratio is defined as the total thrust generated by the four rotors divided by the UAV’s weight, which directly reflects its maximum linear acceleration. The thrust produced by each rotor is proportional to the square of its rotational speed: \begin{equation}
\label{eq1}
T = c_\mathrm{T} \sum_{i=1}^{4} \omega_i^2,
\end{equation} here, $\omega_i$ denotes the rotational speed of rotor $i$, and $c_\mathrm{T}$ is the thrust coefficient determined by the propeller parameters. The maximum thrust-to-weight ratio is given by: \begin{equation}
\label{eq2}
\mathrm{TWR}_{\max}=\frac{T_{\max}}{mg}=\frac{c_\mathrm{T}\sum_{i=1}^4\omega_{i,\max}^2}{mg}
\end{equation} $\omega_{i,\max}$ denotes the maximum achievable rotor speed of motor $i$ under sustained operation.

\begin{figure}[!h]
\centering
\subfloat[“Plus” layout]{\includegraphics[width=1.5in]{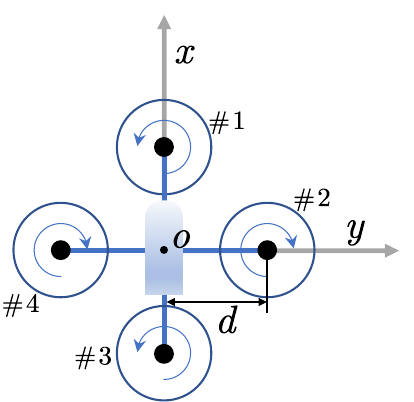}
\label{plus layout}}
\hfil
\subfloat[“Cross” layout]{\includegraphics[width=1.5in]{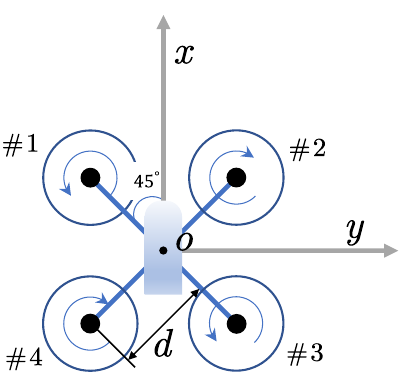}
\label{cross layout}}
\hfil
\caption{{\bf{Comparison of "plus" and "cross" quadrotor layouts.}} This figure illustrates the two common motor configurations: (a) the "plus" layout, where motor arms align with the body's $x$ and $y$ axes, and (b) the "cross" layout, where the arms are rotated by $45^\circ$.}
\label{“plus” layout and “cross” layout}
\end{figure}

{\bf{Maximum Angular Acceleration (Three Axes):}} During navigation, a UAV must also adjust its orientation to change the direction of its thrust vector. This is achieved by generating torques around the three body axes. The reaction torque from each rotor is expressed as: \begin{equation}
\label{eq3}
M_i=c_\mathrm{M}\omega_i^2,
\end{equation} here, $c_\mathrm{M}$ is the torque coefficient of the quadrotor, which is determined by the geometric and aerodynamic properties of the propellers. During rotational motion around the three principal axes, the rotor speeds $\omega_i$ determine the control torque vector $\boldsymbol{\tau} = [\tau_x, \tau_y, \tau_z]^{\mathrm{T}}$. It is important to note that the method for computing these torques varies depending on the quadrotor’s structural configuration, such as the “plus”  or “cross” layout~\cite{ali2022comparing}, as illustrated in \Cref{“plus” layout and “cross” layout}.

{\bf{For a quadrotor with a “plus” layout:}} \begin{equation}
\label{eq4}
\left.\left\{\begin{array}{l}\tau_x=dc_\mathrm{T}(-\omega_2^2+\omega_4^2)\\\tau_y=dc_\mathrm{T}(\omega_1^2-\omega_3^2)\\\tau_z=c_\mathrm{M}(\omega_1^2-\omega_2^2+\omega_3^2-\omega_4^2)\end{array}\right.\right.
\end{equation}

{\bf{For a quadrotor with an “cross” layout:}} \begin{equation}
\label{eq5}
\left.\left\{\begin{array}{l}\tau_x=dc_\mathrm{T}(\frac{\sqrt{2}}{2}\omega_1^2-\frac{\sqrt{2}}{2}\omega_2^2-\frac{\sqrt{2}}{2}\omega_3^2+\frac{\sqrt{2}}{2}\omega_4^2)\\\tau_y=dc_\mathrm{T}(\frac{\sqrt{2}}{2}\omega_1^2+\frac{\sqrt{2}}{2}\omega_2^2-\frac{\sqrt{2}}{2}\omega_3^2-\frac{\sqrt{2}}{2}\omega_4^2)\\\tau_z=c_\mathrm{M}(\omega_1^2-\omega_2^2+\omega_3^2-\omega_4^2)\end{array}\right.\right.
\end{equation}

In the above equation, $d \in \mathbb{R}^+$ represents the distance from the center of the quadrotor body to any of its motors. This value corresponds to the length of the moment arm during rotational motion around the roll or pitch axis.

By modeling the quadrotor as a rigid body, its moment of inertia can be expressed according to the principles of rigid-body dynamics: \begin{equation}
\label{eq6}
\mathbf{J}=\begin{bmatrix}J_{xx}&-J_{xy}&-J_{xz}\\-J_{yx}&J_{yy}&-J_{yz}\\-J_{zx}&-J_{zy}&J_{zz}\end{bmatrix},
\end{equation} here, the inertia tensor satisfies $J_{xy}=J_{yx}$, $J_{xz}=J_{zx}$, and $J_{yz}=J_{zy}$. For a standard quadrotor with center-symmetric geometry, all off-diagonal inertia terms vanish: \begin{equation}
\label{eq7}
J_{xy}=J_{yx}=J_{xz}=J_{zx}=J_{yz}=J_{zy}=0,
\end{equation} hence the tensor simplifies to: \begin{equation}
\label{eq8}
\mathbf{J}=\begin{bmatrix}J_{xx}&0&0\\0&J_{yy}&0\\0&0&J_{zz}\end{bmatrix},
\end{equation} here, $J_{xx},J_{yy},J_{zz}\in{\mathbb{R}^+}$ are the principal moments of inertia. Consequently, the quadrotor’s maximum angular acceleration can be written as: \begin{equation}
\label{eq9}
\begin{aligned}
    \boldsymbol{\alpha}_\mathbf{max} &= \begin{bmatrix}\alpha_{x,\mathrm{max}}\\ \alpha_{y,\mathrm{max}}\\ \alpha_{z,\mathrm{max}} \\\end{bmatrix}=\mathbf{J}^{-1}{\boldsymbol{\tau}_\mathbf{max}} \\
      &= \begin{bmatrix}\frac{1}{J_{xx}} & 0 & 0\\ 0 & \frac{1}{J_{yy}} & 0\\ 0 & 0 & \frac{1}{J_{zz}}\\ \end{bmatrix}\begin{bmatrix}\tau_{x,\mathrm{max}}\\ \tau_{y,\mathrm{max}}\\ \tau_{z,\mathrm{max}} \\\end{bmatrix} \\
      &= \begin{bmatrix}\frac{\tau_{x,\mathrm{max}}}{J_{xx}}\\ \frac{\tau_{y,\mathrm{max}}}{J_{yy}}\\ \frac{\tau_{z,\mathrm{max}}}{J_{zz}} \\\end{bmatrix}
\end{aligned},
\end{equation} FLYINGTRUST evaluates a quadrotor’s dynamic performance using the maximum thrust-to-weight ratio and the maximum angular acceleration along the three axes, which respectively capture its peak thrust generation capability and its ability to rapidly reorient its thrust vector. \begin{equation}
\label{eq10}
\mathbf{P}=\begin{bmatrix}\mathrm{TWR}_{\max}\\\alpha_{x,\mathrm{max}}\\\alpha_{y,\mathrm{max}}\\\alpha_{z,\mathrm{max}}\end{bmatrix}
\end{equation}

Because a quadrotor is typically designed to be center-symmetric, its principal moments of inertia about the x and y axes are often treated as equal ($J_{xx} \approx J_{yy}$). However, the inclusion of onboard components, such as a fixed forward-facing camera, introduces a slight mass asymmetry. Strictly speaking, this results in a minor difference between the maximum roll acceleration ($\alpha_{x,\mathrm{max}}$) and pitch acceleration ($\alpha_{y,\mathrm{max}}$). For this benchmark, we justify the use of a unified horizontal angular acceleration metric $\alpha_{xy,\mathrm{max}}$ for two primary reasons:

\begin{itemize}
  \item {\bf{Dominance of Roll in Navigation:}} In many autonomous navigation scenarios, especially those involving lateral obstacle avoidance at speed, roll maneuvers are the most frequent and dynamically demanding. The vehicle's ability to quickly roll and redirect its thrust sideways is often more critical than its pitching capability for acceleration or deceleration. Therefore, the roll performance ($\alpha_{x,\mathrm{max}}$) serves as a practical and representative limit for horizontal agility. 
  \item  {\bf{Negligible Asymmetry Impact:}} For the majority of small to medium-sized quadrotors considered in our dataset, the mass of the camera and other directional sensors is a small fraction of the total vehicle mass. This results in a minimal difference between $J_{xx}$ and $J_{yy}$, making the discrepancy between their corresponding maximum angular accelerations negligible for a high-level performance benchmark.
\end{itemize}

Given these practical considerations, this paper uses the dominant roll acceleration value $\alpha_{x,\mathrm{max}}$ to represent the combined horizontal agility metric $\alpha_{xy,\mathrm{max}}$. This simplifies the kinodynamic model effectively without a significant loss of fidelity for the evaluation of forward-flight navigation tasks. Given this simplification and the symmetric motor configuration described in \Cref{eq4,eq5}, we proceed by using a single horizontal agility metric. Hence, FLYINGTRUST’s characterization of kinodynamics can be reduced to: \begin{equation}
\label{eq11}
\mathbf{P}=\begin{bmatrix}\mathrm{TWR}_{\max}\\\alpha_{xy,\mathrm{max}}\\\alpha_{z,\mathrm{max}}\end{bmatrix}
\end{equation}

This study surveyed the propeller, motor, and frame parameters of several commercial off-the-shelf quadrotors and custom research platforms, and used these data to estimate each platform’s kinodynamics performance metrics~\cite{shi2017practical}. Based on the proprietary platforms used in EGO-Planner, Fast-Planner and Agilicious, we synthesized 16 virtual UAVs. These models were generated by systematically scaling key parameters, such as vehicle mass, arm length $d$, and maximum propeller speed within the documented design space. From these scaled values, we then recalculated the associated moments of inertia ($J$) and the final kinodynamic performance metrics ($\mathrm{TWR}_{\max}$, $\alpha_{xy,\mathrm{max}}$, $\alpha_{z,\mathrm{max}}$) to ensure physical consistency. Each virtual configuration was subjected to basic feasibility checks, for example, the consistency of mass and inertia scaling and motor/propeller limits, and validated in simulation for grossly infeasible behaviour~\cite{mellinger2012trajectory}. The resulting benchmark therefore combines 18 real platforms with 18 virtual models to form the FLYINGTRUST kinodynamics dataset.

\begin{table}[!ht]
\centering
\caption{Summary of Performance Parameters for Real and Virtual UAV Platforms\label{tab1}}
\resizebox{\linewidth}{!}{
\begin{tabular}{@{}ll*{3}{r}@{}}
\toprule
\bf{Category} & \bf{UAV Platform} & $\mathbf{\mathrm{TWR}}_{\max}$ & $\boldsymbol{\alpha}_{xy,\mathrm{max}}(\mathrm{rad/s^2})$ & $\boldsymbol{\alpha}_{z,\mathrm{max}}(\mathrm{rad/s^2})$ \\
\midrule
\multirow{18}{*}{\makecell[c]{\bfseries Real Vehicles}} & 0.60kg-EMAX & 2.2 & 114.7 & 8.4 \\
                     & 0.895kg-DJI & 3.3 & 107.9 & 14.1 \\
                     & 0.90kg-DJI & 3.0 & 139.2 & 10.5 \\
                     & 1.00kg-SunnySky & 6.0 & 227.3 & 13.9 \\
                     & 1.20kg-JFRC & 1.4 & 84.6 & 7.2 \\
                     & 1.40kg-EMAX & 2.5 & 94.1 & 6.0 \\
                     & 1.50kg-DJI & 1.8 & 85.8 & 5.7 \\
                     & 1.80kg-SunnySky & 2.3 & 127.8 & 7.7 \\
                     & 2.00kg-T-MOTOR & 1.4 & 55.6 & 3.3 \\
                     & 2.50kg-HLY & 1.5 & 65.1 & 4.6 \\
                     & 2.80kg-T-MOTOR & 2.5 & 79.9 & 4.6 \\
                     & 3.00kg-T-MOTOR & 2.2 & 112.0 & 7.6 \\
                     & 3.50kg-SunnySky & 1.4 & 116.2 & 9.5 \\
                     & 3.80kg-T-MOTOR & 1.4 & 63.6 & 4.1 \\
                     & 4.00kg-SunnySky & 1.9 & 83.7 & 5.0 \\
                     & 4.50kg-T-MOTOR & 1.8 & 95.7 & 6.8 \\
                     & 4.91kg-DJI & 2.5 & 69.6 & 6.9 \\
                     & 5.45kg-JFRC & 2.6 & 75.8 & 3.3 \\
% \addlinespace
\midrule
\multirow{18}{*}{\makecell[c]{\bfseries Virtual Vehicles}} & 0.55kg-UAV 1 & 3.6 & 1383.7 & 69.2 \\
                     & 0.68kg-Agile Autonomy DIY~\cite{loquercio2021learning} & 3.0 & 171.4 & 17.4 \\ 
                     & 0.75kg-UAV 2 & 4.2 & 1467.0 & 73.3 \\
                     & 0.85kg-UAV 3 & 3.2 & 1052.7 & 52.6 \\
                     & 0.98kg-EGO Planner DIY~\cite{zhou2020ego} & 4.6 & 1083.3 & 57.7 \\
                     & 1.05kg-UAV 4 & 3.5 & 950.7 & 47.5 \\
                     & 1.20kg-UAV 5 & 4.0 & 1164.8 & 58.2 \\
                     & 1.50kg-UAV 6 & 3.8 & 931.1 & 46.5 \\
                     & 1.80kg-UAV 7 & 3.8 & 969.3 & 48.5 \\
                     & 2.00kg-UAV 8 & 3.2 & 712.9 & 35.6 \\
                     & 2.50kg-UAV 9 & 3.0 & 692.8 & 34.6 \\
                     & 2.80kg-UAV 10 & 3.4 & 694.5 & 34.7 \\
                     & 3.00kg-UAV 11 & 3.3 & 697.4 & 34.9 \\
                     & 3.50kg-UAV 12 & 3.1 & 584.4 & 29.2 \\
                     & 4.20kg-UAV 13 & 2.8 & 553.4 & 27.7 \\
                     & 4.50kg-UAV 14 & 2.9 & 507.8 & 25.4 \\
                     & 4.80kg-UAV 15 & 3.6 & 587.3 & 29.4 \\
                     & 5.00kg-UAV 16 & 3.5 & 631.0 & 31.5 \\                   
\bottomrule
\end{tabular}
}
\end{table}

\begin{figure}[!h]
\centering
\subfloat[$\mathrm{TWR}_{\max}$ vs mass]{\includegraphics[width=3.4in]{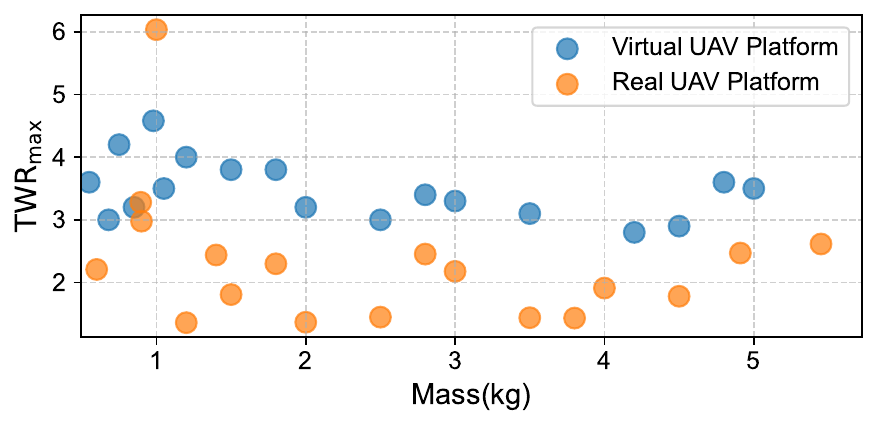}
\label{kinodynamic_1}}
\hfil

\subfloat[$\alpha_{xy,\mathrm{max}}$ vs mass]{\includegraphics[width=3.4in]{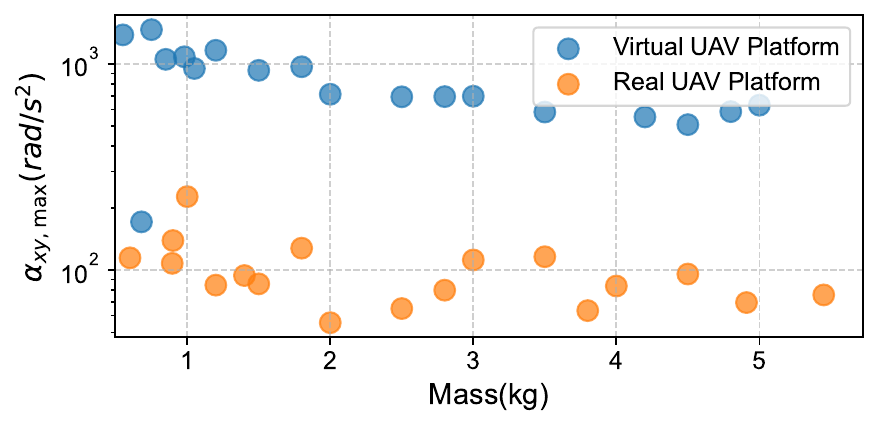}
\label{kinodynamic_2}}
\hfil

\subfloat[$\alpha_{z,\mathrm{max}}$ vs mass]{\includegraphics[width=3.4in]{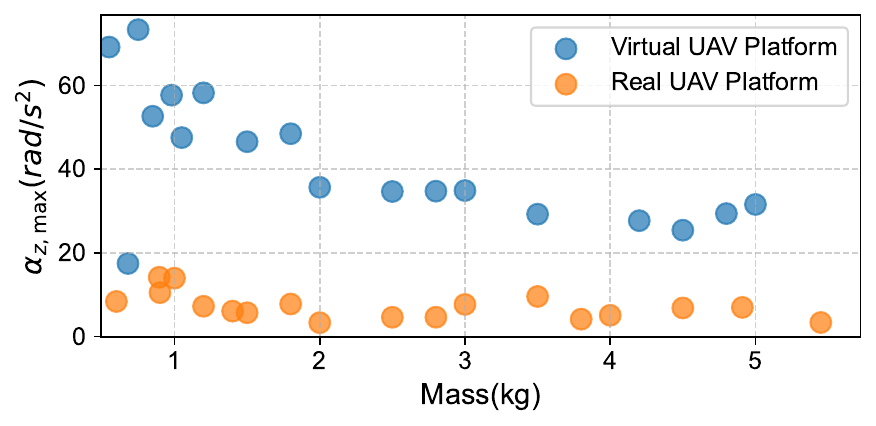}
\label{kinodynamic_3}}
\hfil
\caption{{\bf{UAV kinodynamic properties in FLYINGTRUST.}} Scatter plots showing the relationship between platform mass and the three kinodynamic metrics: (a) maximum thrust-to-weight ratio ($\mathrm{TWR}_{\max}$), (b) maximum horizontal angular acceleration ($\alpha_{xy,\mathrm{max}}$, log scale), and (c) maximum yaw angular acceleration ($\alpha_{z,\mathrm{max}}$). Virtual UAVs are marked in blue; real UAVs are marked in orange.}
\label{kinodynamic}
\end{figure}

As illustrated in \Cref{tab1} and \Cref{kinodynamic}, FLYINGTRUST focuses on small to medium sized quadrotors, covering a weight range from $0.5~\mathrm{kg}$ to $5~\mathrm{kg}$. In the real-world subset, the mean thrust-to-weight ratio is $2.30$, the mean maximum angular acceleration about the horizontal axes (x,y) is $99.92~\mathrm{rad/s^2}$, and about the yaw axis (z) is $7.17~\mathrm{rad/s^2}$. For the virtual models, the mean thrust-to-weight ratio is $3.47$, the mean maximum angular acceleration about the horizontal axes is $824.20~\mathrm{rad/s^2}$, and about the yaw axis is $41.88~\mathrm{rad/s^2}$. These results confirm that most commercial platforms prioritize flight smoothness over agility, resulting in substantially lower dynamic performance compared with idealized UAVs. Moreover, existing navigation algorithms typically assume high thrust-to-weight ratios and angular accelerations during their design, training, and validation stages.

\subsection{Navigation scenarios}

\begin{figure*}[!ht]
\centering
\subfloat[Forest]{\includegraphics[width=2.5in]{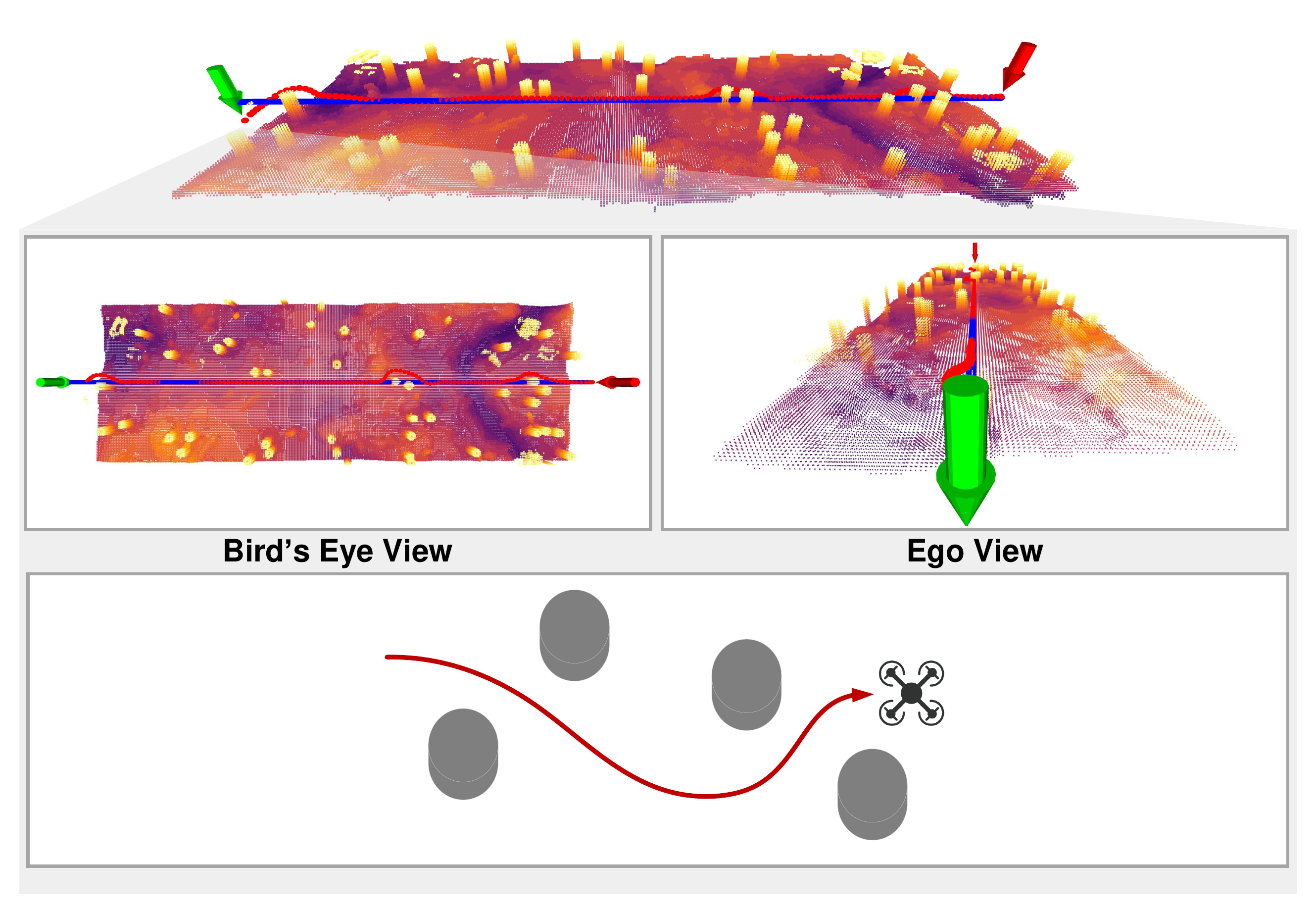}%
\label{fig_forest}}
\hfil
\subfloat[Urban]{\includegraphics[width=2.5in]{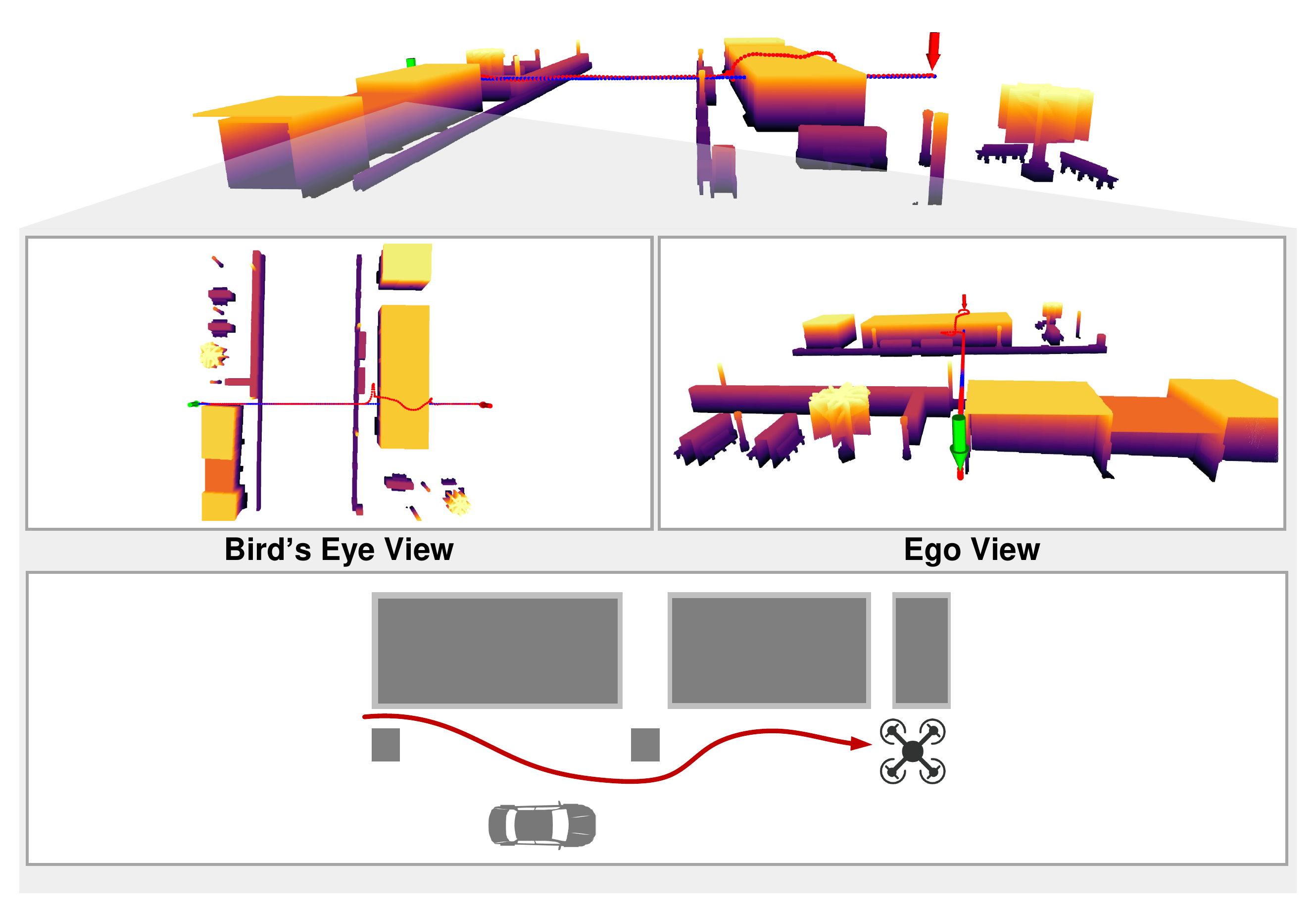}%
\label{fig_urban}}
\hfil

\subfloat[Random-Angle Cylinder]{\includegraphics[width=2.5in]{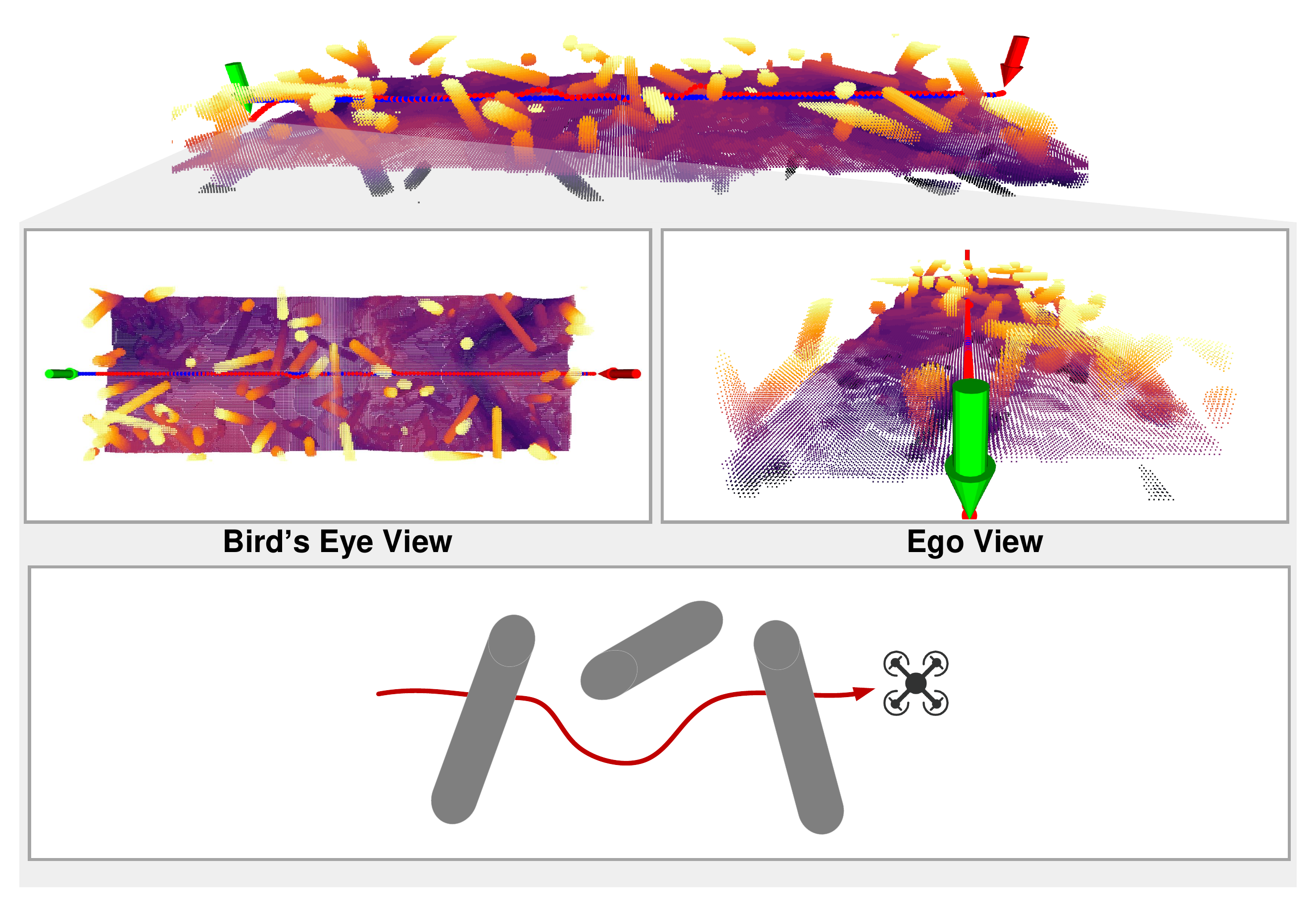}%
\label{fig_random_cylinder}}
\hfil
\subfloat[Narrow-Gap]{\includegraphics[width=2.5in]{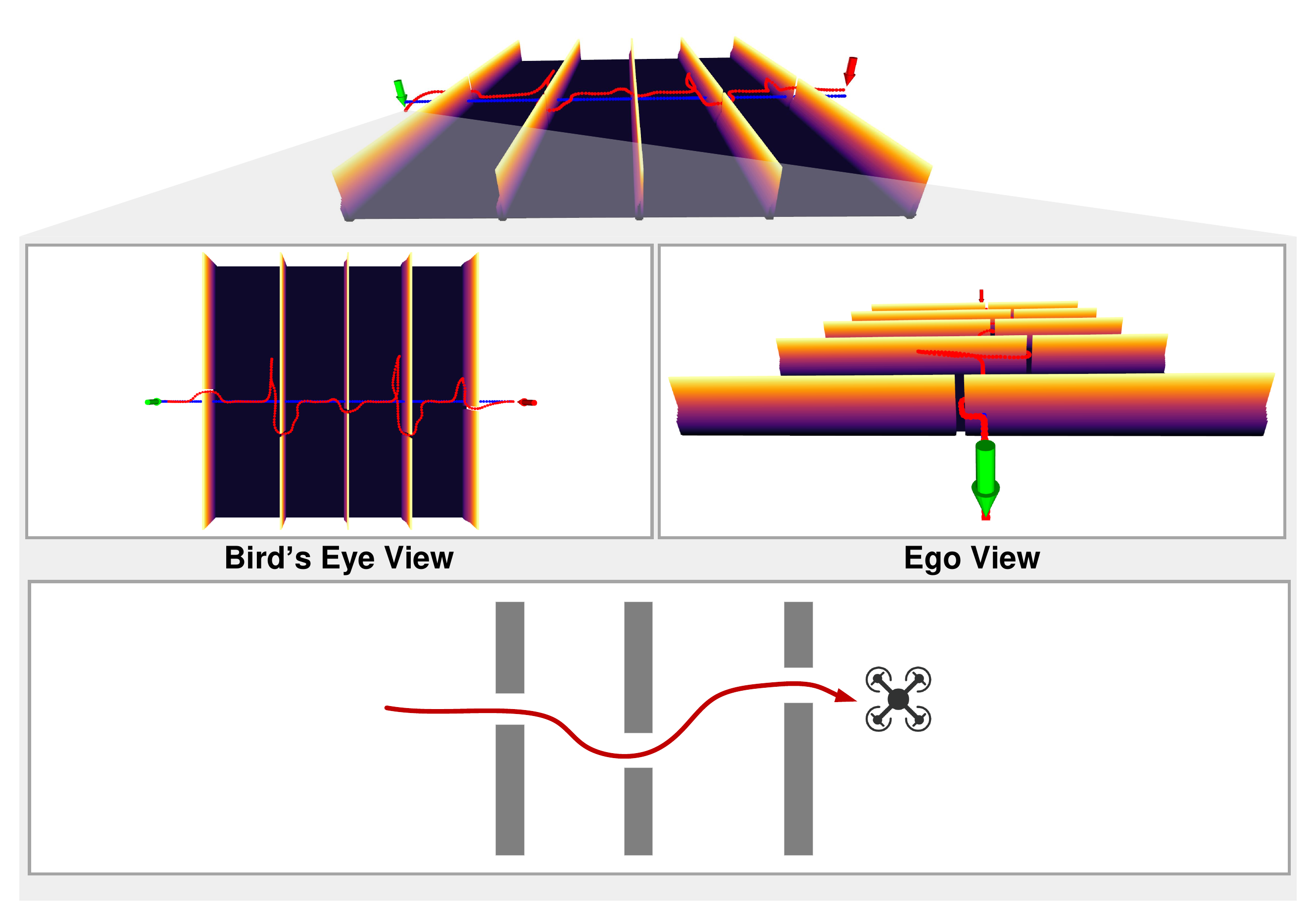}%
\label{fig_narrow_gap}}
\hfil

\subfloat[Sudden-Drop]{\includegraphics[width=2.5in]{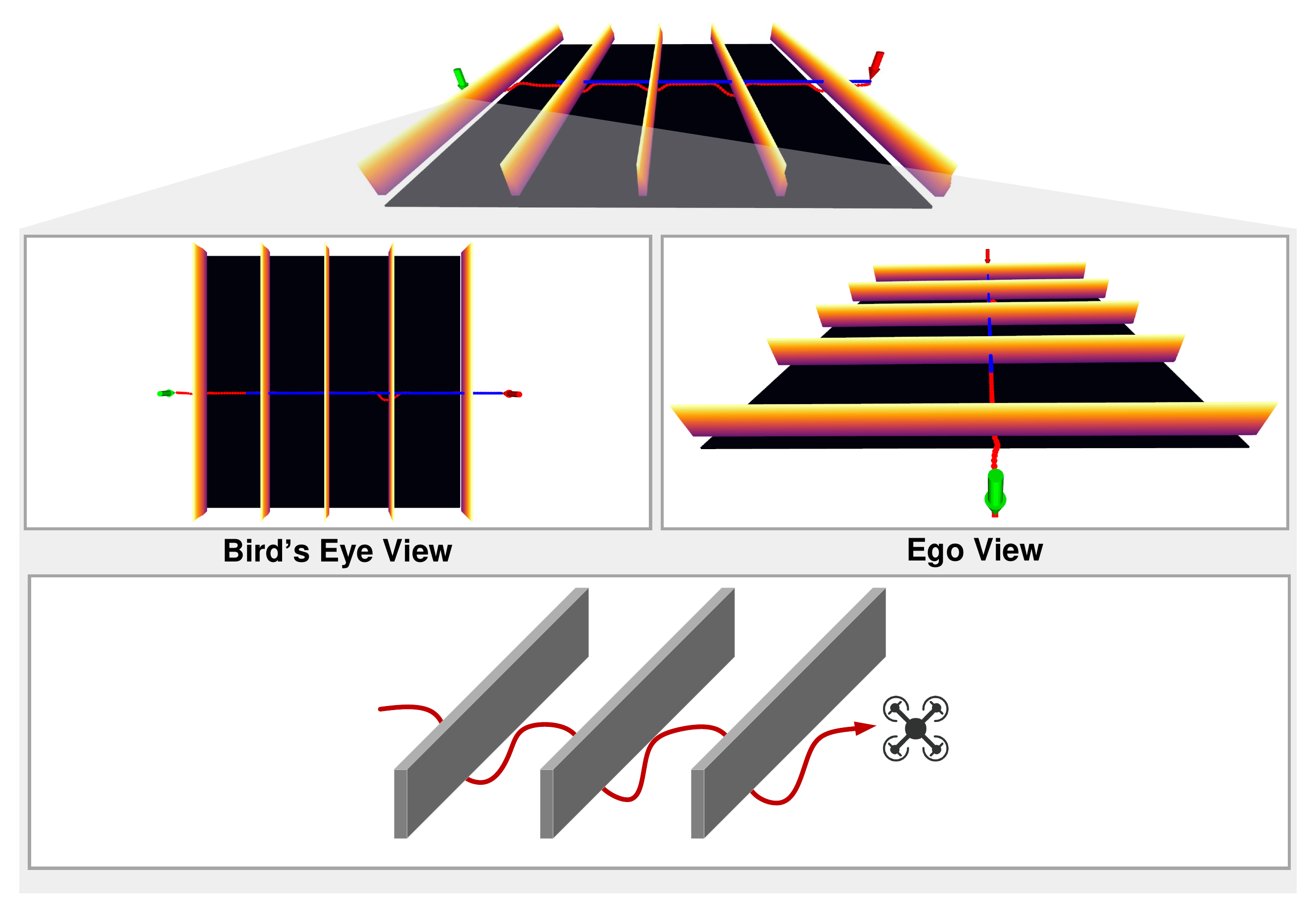}%
\label{fig_sudden_drop}}
\hfil
\subfloat[Maze]{\includegraphics[width=2.5in]{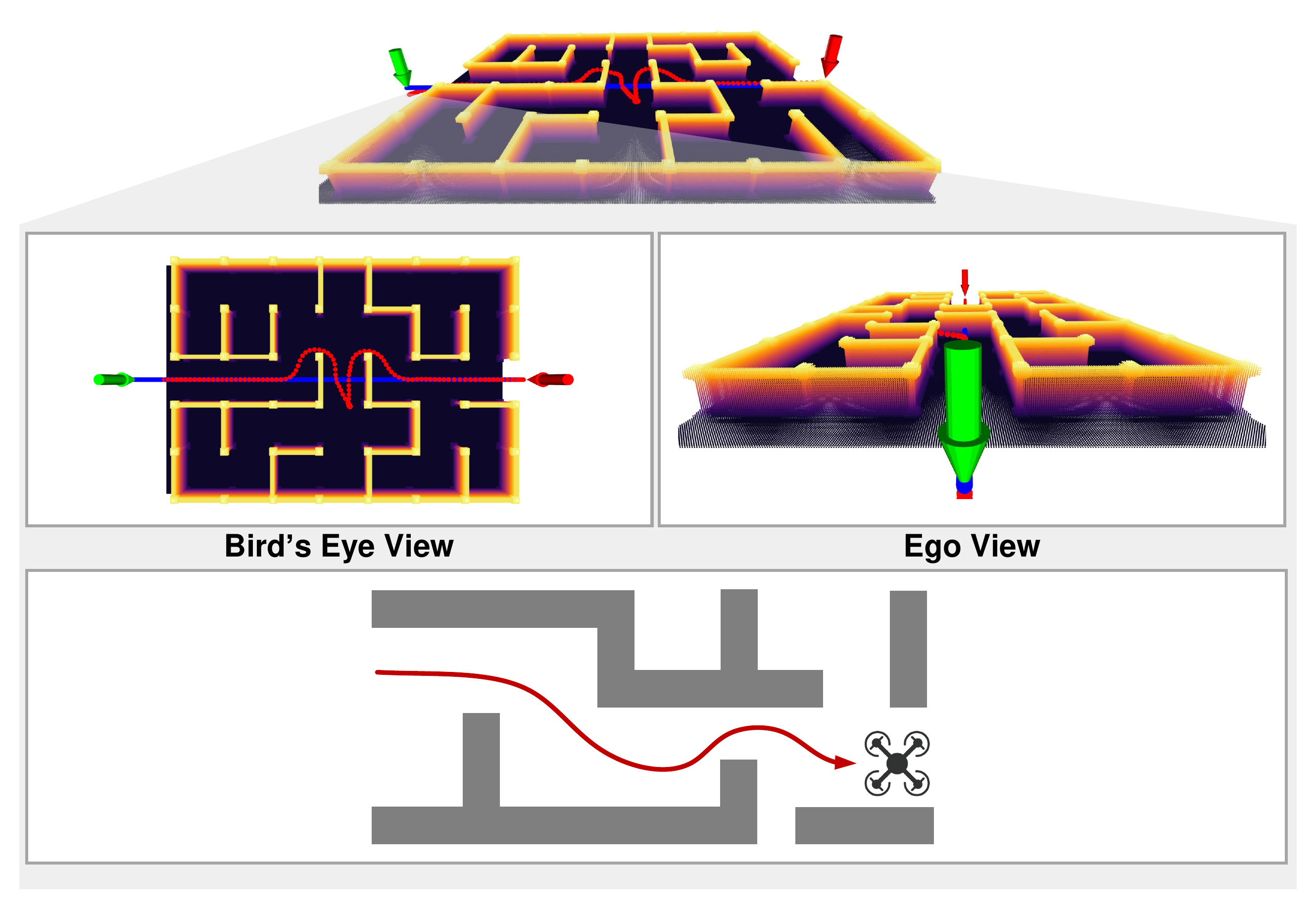}%
\label{fig_maze}}
\hfil

\subfloat[Random Perlin-Noise]{\includegraphics[width=2.5in]{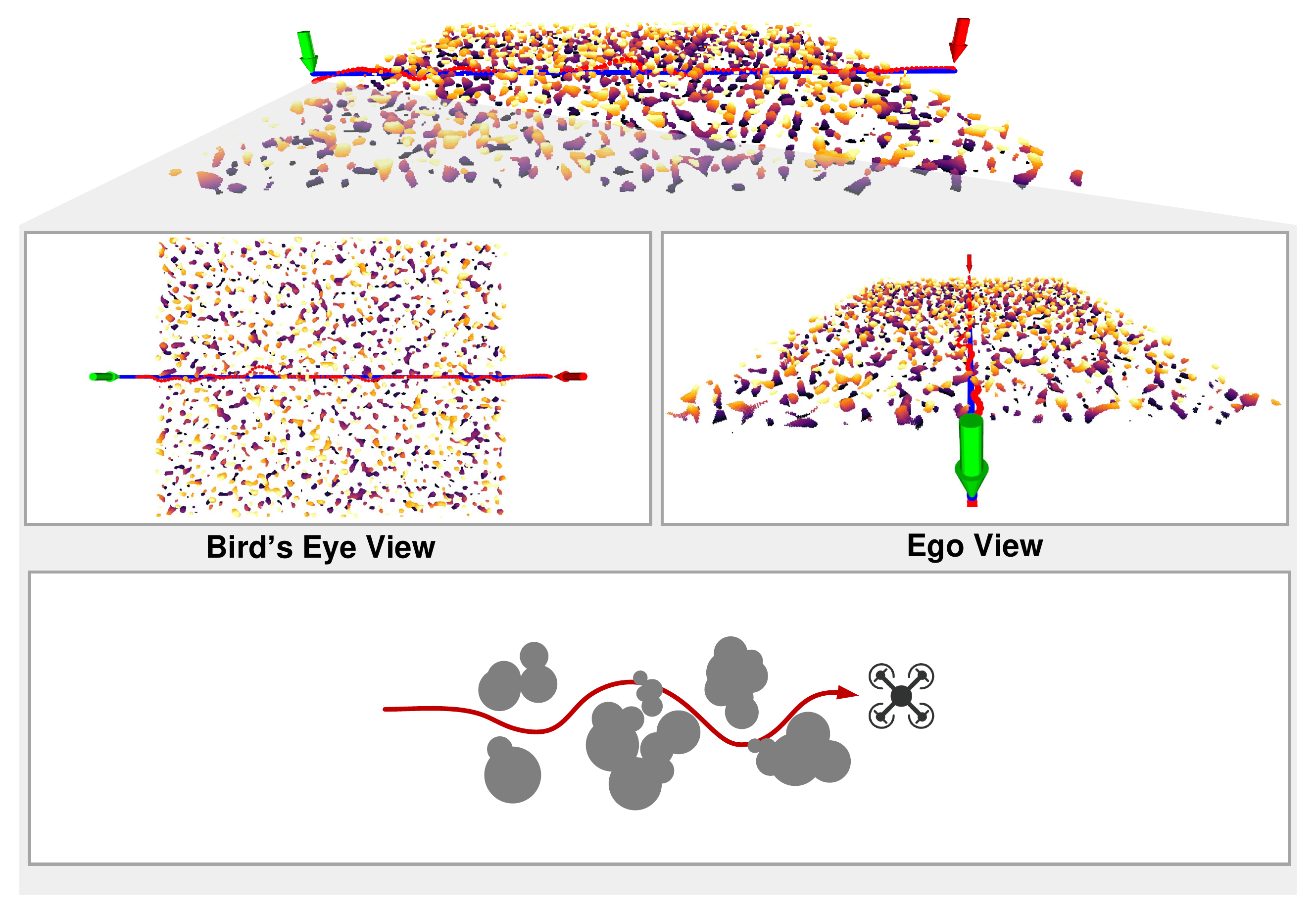}%
\label{fig_perlin_noise}}
\hfil
\caption{{\bf{Navigation scenarios used in FLYINGTRUST.}} The seven scenarios are grouped into (a-c) Classic Navigation Scenarios (Forest, Urban, Random-Angle Cylinder) and (d-g) Theoretical Navigation Scenarios (Narrow-Gap, Sudden-Drop, Maze, Random Perlin-Noise). In each subplot, the green arrow marks the start pose, the red arrow marks the goal, the blue line is the straight-line reference path, and the red curve is an example collision-free trajectory.}
\label{navigation scenarios}
\end{figure*}

FLYINGTRUST defines a suite of challenging navigation scenarios to evaluate the robustness of visual navigation algorithms across diverse environments. Its scenarios are categorized into Classic Navigation Scenarios and Theoretical Navigation Scenarios. Classic scenarios represent real-world, commonly used test environments, assessing algorithm performance under typical conditions; theoretical scenarios are specially crafted extreme cases designed to probe specific algorithmic capabilities and reveal potential weaknesses. The FLYINGTRUST dataset is provided in .PCD, .PLY, and Gazebo-compatible .dae formats, seamlessly integrating with existing quadrotor simulation platforms. Note that at the time of this study the Random Perlin-Noise scenario has not been successfully migrated to Gazebo; where this scenario is absent we mark the affected algorithm entries and apply the missing-data policy described in \Cref{scoring_method}.

\subsubsection{Classic Navigation Scenarios}
As illustrated in \Cref{fig_forest,fig_urban,fig_random_cylinder}, classic navigation scenarios consist of forest, urban, and random-angle cylinder obstacle environments. These scenarios are commonly used in visual navigation research to assess algorithm performance under typical, real-world conditions. Therefore, we expect the evaluated methods to adapt well and achieve high success rates in these classic tests.

\begin{itemize}
  \item {\bf{Forest}}: FLYINGTRUST leverages Unity~\cite{juliani2018unity} to create a canonical forest scenario measuring $40~\mathrm{m}$ in width and $60~\mathrm{m}$ in length, with a drone flight ceiling of $3~\mathrm{m}$. Using a tree density of $\delta=1/49$, we randomly generated ten distinct obstacle configurations to form the forest navigation dataset. These scenarios are subsequently reconstructed and validated in the Gazebo simulator~\cite{koenig2004design}. Such settings are widely adopted in UAV studies to benchmark perception and obstacle avoidance in unstructured outdoor environments~\cite{loquercio2021learning,zhou2020ego}. The primary challenge lies in reliably detecting narrow gaps between trees while maintaining high-speed maneuvering. 
  \item {\bf{Urban}}: FLYINGTRUST uses Unity to create a canonical urban environment measuring $60~\mathrm{m}$ in width and $60~\mathrm{m}$ in length, with a drone flight ceiling of $10~\mathrm{m}$. Ten obstacle configurations were randomly sampled from the Unity urban scenario project to form the urban navigation dataset, which was then reconstructed and validated in the Gazebo simulator. The urban scenario introduces structured but cluttered geometry with vertical obstacles resembling buildings and walls. Compared to forests, urban layouts pose challenges for visual localization (e.g., repeated textures, sharp corners) and can induce failure modes such as drift or premature collision.
  \item {\bf{Random-Angle Cylinder}}: FLYINGTRUST builds a randomized, tilted-cylinder obstacle field in Unity, covering an area $40~\mathrm{m}$ wide by $60~\mathrm{m}$ long with a flight ceiling of $3~\mathrm{m}$. Using a cylinder density of $\delta=1/36$, radii ranging from $0.25~\mathrm{m}$ to $0.5~\mathrm{m}$, and tilt angles between $0^\circ$ and $180^\circ$, ten distinct obstacle configurations were generated to constitute the randomly oriented cylinder obstacle dataset. These scenarios are subsequently reconstructed and validated in Gazebo. Randomly tilted cylinders create unpredictable obstacle orientations, breaking the symmetry of standard upright obstacles. This design challenges UAV algorithms to handle highly variable geometry and perform adaptive lateral maneuvers~\cite{ren2025safety}. Success in this environment indicates robustness to non-canonical obstacle distributions.
\end{itemize}

\subsubsection{Theoretical Navigation Scenarios}
As illustrated in \Cref{fig_narrow_gap,fig_sudden_drop,fig_maze,fig_perlin_noise}, theoretical navigation problem scenarios include narrow-gap, sudden-drop, maze, and random Perlin-noise environments. Crafted from failure cases observed in extensive engineering trials, these extreme scenarios target specific challenges—lateral large-yaw maneuvers, vertical large-pitch maneuvers, complex path exploration, and high-density random obstacles. Given their elevated difficulty, we anticipate lower success rates in these theoretical tests compared to the classic scenarios.

\begin{itemize}
  \item {\bf{Narrow-Gap}}: FLYINGTRUST constructs a $50~\mathrm{m}$ by $50~\mathrm{m}$ narrow-gap environment in Unity with a flight ceiling of $4~\mathrm{m}$. Gap widths range from $0.85~\mathrm{m}$ to $0.9~\mathrm{m}$. Ten obstacle configurations are generated at random, with the number of gaps increasing from Scenario 1 to Scenario 10, forming the narrow-gap dataset. These scenarios are reconstructed in Gazebo. With openings as narrow as $0.85~\mathrm{m}$, this environment explicitly tests a UAV’s capability to execute large-yaw maneuvers for lateral gap navigation. Similar settings are often used in agile flight benchmarks, where insufficient yaw agility or inaccurate perception leads to frequent crashes~\cite{falanga2020dynamic}.
  \item {\bf{Sudden-Drop}}: FLYINGTRUST creates a sudden-drop environment in Unity measuring $50~\mathrm{m}$ by $50~\mathrm{m}$ by $14~\mathrm{m}$, where the lowest obstacle point is $1.5~\mathrm{m}$ above ground. The UAV starts at $2.5~\mathrm{m}$ altitude with a ceiling of $4~\mathrm{m}$. Ten random obstacle configurations are generated, with obstacle count rising from Scenario 1 to Scenario 10, forming the sudden-drop dataset. These scenarios are reproduced in Gazebo to evaluate whether an algorithm can handle rapid pitch adjustments and altitude control when encountering sharp terrain discontinuities. UAVs with limited pitch authority often fail by colliding with obstacles after delayed descent response~\cite{torrente2021data}.
  \item {\bf{Maze}}: FLYINGTRUST uses a maze generator in Unity to design a $25~\mathrm{m}$ by $40~\mathrm{m}$ labyrinth, limiting flight height to $2~\mathrm{m}$. Ten random maze layouts form the maze dataset, which is reconstructed in Gazebo. The maze environment demands deliberate exploration and decision-making under partial observability. It reflects real-world scenarios such as indoor navigation or search-and-rescue, where dead ends force replanning and robust memory integration.
  \item {\bf{Random Perlin-Noise}}: FLYINGTRUST designs a $40~\mathrm{m}$ by $50~\mathrm{m}$ random Perlin noise environment with a flight ceiling of $4~\mathrm{m}$, using a noise frequency of $0.05$ and voxel fill rate of $0.03$. Ten random obstacle configurations constitute the Perlin noise dataset. Migrating this scenario to Gazebo proved challenging, and a reconstruction has not yet been achieved. By generating highly irregular obstacle fields through Perlin noise, this scenario produces extreme clutter and visual aliasing. It pushes algorithms to their limits in perception, mapping, and global planning.
\end{itemize}

\subsection{Scoring Method}
\label{scoring_method}
After conducting a series of navigation success rate tests, we further compare and rank different visual UAV navigation algorithms by introducing a composite scoring system based on scenario and platform weighting. This method draws inspiration from composite performance indices and multi-criteria decision making (MCDM)~\cite{triantaphyllou2000multi}, and integrates three dimensions: scenario importance, platform importance, and algorithmic stability.

The overall idea is as follows: For each algorithm, the navigation success rates $S_{a,s,m}$ across different scenarios and UAV models are aggregated using weighted averages, where the weights encode scenario and platform importance. To account for consistency, a variance-based penalty is then applied to discourage algorithms whose performance fluctuates strongly across conditions~\cite{zahavy2016ensemble}. The final score $\mathrm{FinalScore}_a$ serves as an intuitive “benchmark score” for comparison.

{\bf{Scenario weights}}: Each scenario type (classic vs. theoretical) is assigned an initial weight $w_s (s \in \mathcal{S})$. After normalization, \begin{equation}
\label{eq12}
W_s=\frac{w_s}{\sum_{s^{\prime}}w_{s^{\prime}}}
\end{equation}

{\bf{Platform weights}}: Each UAV model type (real vs. virtual) is assigned a weight $w_m (m \in \mathcal{M})$. After normalization, \begin{equation}
\label{eq13}
W_m=\frac{w_m}{\sum_{m^{\prime}}w_{m^{\prime}}}
\end{equation}

{\bf{Initial score}}: The weighted mean success rate is computed as \begin{equation}
\label{eq14}
\widehat{\mathrm{Score}}_a=\frac{\sum_{s}\sum_{m}W_s W_m S_{a,s,m}}{\sum_{s}\sum_{m}W_s W_m},
\end{equation} which is scaled to percentage form: \begin{equation}
\label{eq15}
\mathrm{Score}_a = 100 \times \widehat{\mathrm{Score}}_a
\end{equation}

{\bf{Stability penalty}}: To reflect consistency, the weighted variance is defined as \begin{equation}
\label{eq16}
\mathrm{Var}_a=\sum_{s}\sum_{m} W_sW_m(S_{a,s,m}-\widehat{\mathrm{Score}}_a)^2,
\end{equation} and normalized as \begin{equation}
\label{eq17}
\mathrm{Var}^{\mathrm{norm}}_a=\frac{\mathrm{Var}_a}{\max_{a^{\prime}}\mathrm{Var}_{a^{\prime}}}\in[0,1]
\end{equation}

{\bf{Final score}}: Incorporating the stability penalty, the final score is \begin{equation}
\label{eq18}
\mathrm{FinalScore}_a=\mathrm{Score}_a\times\left(1-\beta\cdot \mathrm{Var}_a^{\mathrm{norm}}\right),
\end{equation} where $\beta \in [0,1]$ controls the penalty strength. $\beta =0$ yields a purely average-based score, while $\beta > 0$ emphasizes algorithms with more stable performance across heterogeneous conditions.

For algorithms that lack results for an entire scenario $s_0$, we exclude $s_0$ from the summation and renormalize the remaining scenario weights for that algorithm: \begin{equation}
\label{eq19}
W_s^{\prime}=\frac{W_s}{\sum_{s\in \mathcal{S}\setminus\{s_0\}}W_s},
\end{equation} where $W_s$ denotes the original normalized scenario weight and $W_s^{\prime}$ denotes the renormalized weight after exclusion.

\section{Experiments}
\label{Experiments}
\subsection{Algorithms}
We evaluated several commonly used and widely deployed quadrotor navigation algorithms:

\begin{itemize}
  \item {\bf{EGO-Planner}}~\cite{zhou2020ego}: An ESDF-free local trajectory optimizer that relies on guided paths and anisotropic curve fitting to achieve collision avoidance and smooth trajectories.
  \item {\bf{Fast-Planner}}~\cite{fast-planner}: A two-stage approach that first searches for a dynamically feasible path using discrete motion primitives, then refines it via B-spline optimization to enhance geometric quality and ensure dynamic feasibility, striking a balance between real-time performance and flight quality.
  \item {\bf{Path-Guided PGO}}~\cite{PGO}: A replanning method that samples diverse topological paths to escape local minima, using these routes as priors to guide the optimizer toward higher-quality trajectories in complex scenarios.
  \item {\bf{NavRL}}~\cite{xu2025navrl}: A deep reinforcement learning framework employing PPO and specialized perception modules for static and dynamic obstacles, augmented by a velocity-obstacle safety shield to enable zero-shot sim-to-real transfer and robust obstacle avoidance in dynamic environments. Testing employed the authors’ publicly released pre-trained model, obtained from their GitHub repository~\cite{navrlcode}, without further fine tuning.
  \item {\bf{Agilicious}}~\cite{foehn2022agilicious}: An open-source hardware and software project that provides a modular stack for agile vision-based quadrotor flight, including high-rate control loops, perception interfaces and a simulation ecosystem for development and real-world deployment. Agilicious emphasizes low-level control fidelity and real-time responsiveness, and it is well suited to research on aggressive maneuvers and high-agility platforms.
  \item {\bf{Agile-Autonomy}}~\cite{loquercio2021learning}: A learning-based approach that targets robust, high-speed navigation across diverse real-world environments by combining data-driven policy training with careful sim-to-real engineering. The method demonstrates strong performance on agile flight tasks and is particularly relevant for high-velocity, perception-driven autonomy.
  \item {\bf{Straight-Flight Baseline}}: A constant-velocity direct path from start to goal without obstacle avoidance, used as a baseline to quantify the benefit of active planning methods.
\end{itemize}

Agilicious and Agile-Autonomy are not evaluated here because their simulation and software ecosystems are tightly coupled to specific vehicle configurations, complicating integration with the FLYINGTRUST scenes. Incorporating these systems in a fair and reproducible way would require extensive engineering adaptation beyond the intended scope of this study.

It is important to note that our evaluation compares multiple optimization-based planners against a single, representative learning-based algorithm (NavRL), using its publicly available pre-trained model. Consequently, the observed behaviors and limitations of NavRL may be specific to its architecture and training data, and should not be interpreted as fundamental properties of all learning-based navigation approaches.

For fair comparison, all methods (including the straight-flight baseline) use the same start pose. All runs respect identical flight ceilings, maximum speed limits and acceleration constraints.

\subsection{Benchmarking results}
To evaluate a set of representative quadrotor navigation algorithms, FLYINGTRUST conducted a series of simulations. In these tests, the UAV’s maximum flight speed was capped at $4~\mathrm{m/s}$. The task consisted of point-to-point navigation: The UAV first flew from its takeoff position to a designated start waypoint and then followed the algorithm’s planned path to the goal. A straight line connecting start and goal served as the reference trajectory, with obstacles placed along it. Success was defined by the UAV reaching the goal within $1.5~\mathrm{minutes}$ under the $4~\mathrm{m/s}$ speed limit and remaining stably within a $2~\mathrm{m}$ radius of the target. Failure was recorded if the UAV collided with any obstacle, if its final stable position lay outside the $2~\mathrm{m}$ radius, or if it failed to plan a viable path to the goal within $1.5~\mathrm{minutes}$.

\begin{figure*}[!ht]
\centering
\includegraphics[width=6.5in]{"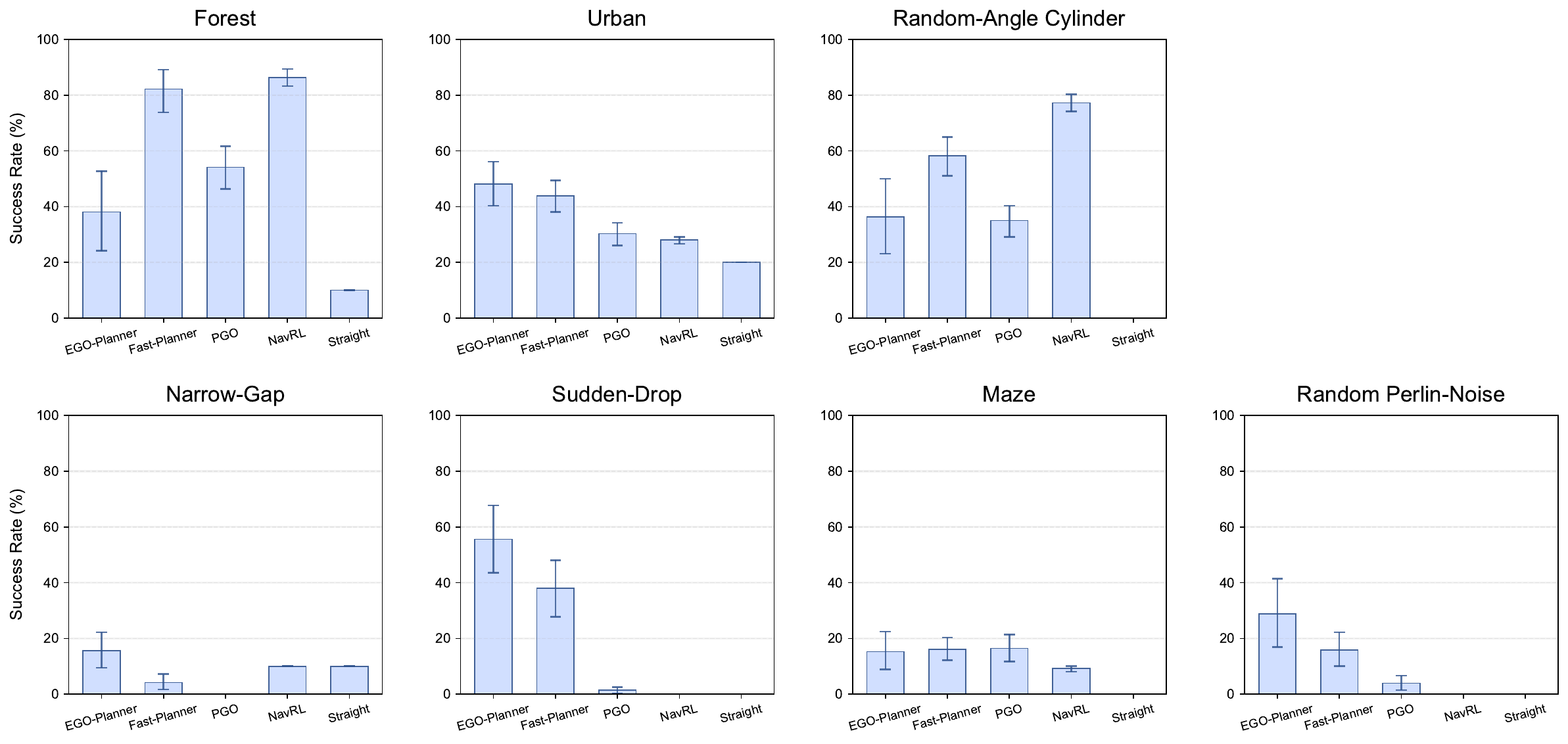"}
\caption{{\bf{Benchmark results by navigation scenario.}} The mean navigation success rate for each algorithm, averaged across all 36 UAV platforms, is shown for each of the seven scenarios. Error bars indicate 95\% bootstrap confidence intervals.}
\label{different_scenarios}
\end{figure*}

\begin{figure*}[!ht]
\centering
\subfloat[Success rate vs $\mathrm{TWR}_{\max}$]{\includegraphics[width=2.3in]{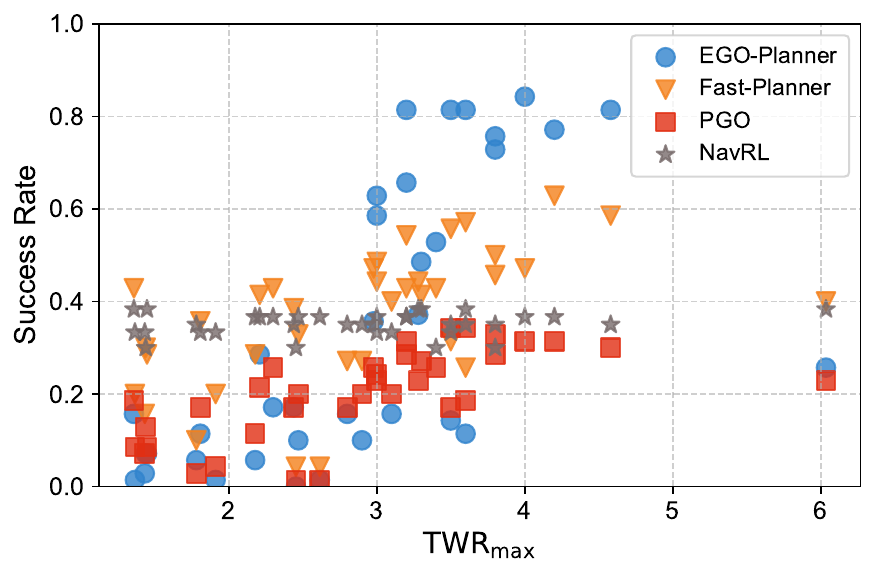}
\label{success_rate_analysis_1}}
\hfil
\subfloat[Success rate vs $\alpha_{xy,\mathrm{max}}$]{\includegraphics[width=2.3in]{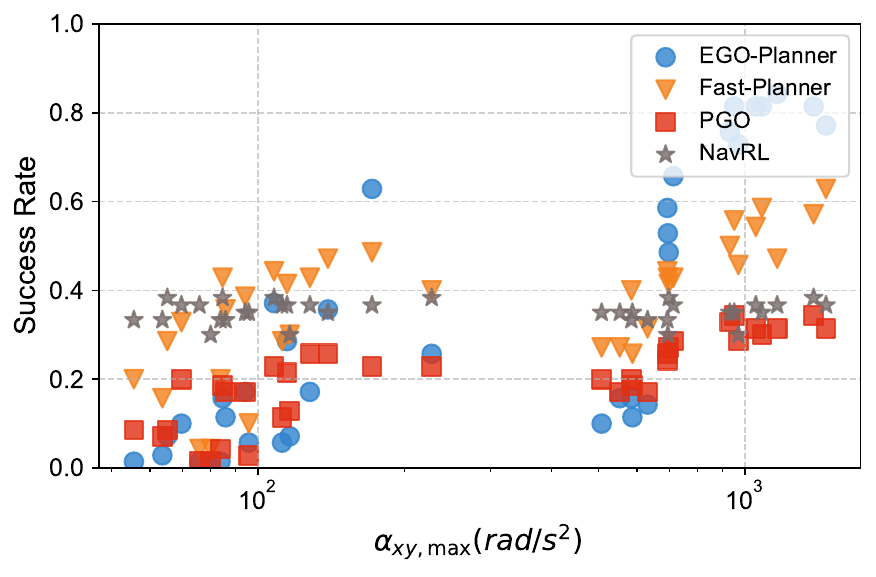}
\label{success_rate_analysis_2}}
\hfil
\subfloat[Success rate vs $\alpha_{z,\mathrm{max}}$]{\includegraphics[width=2.3in]{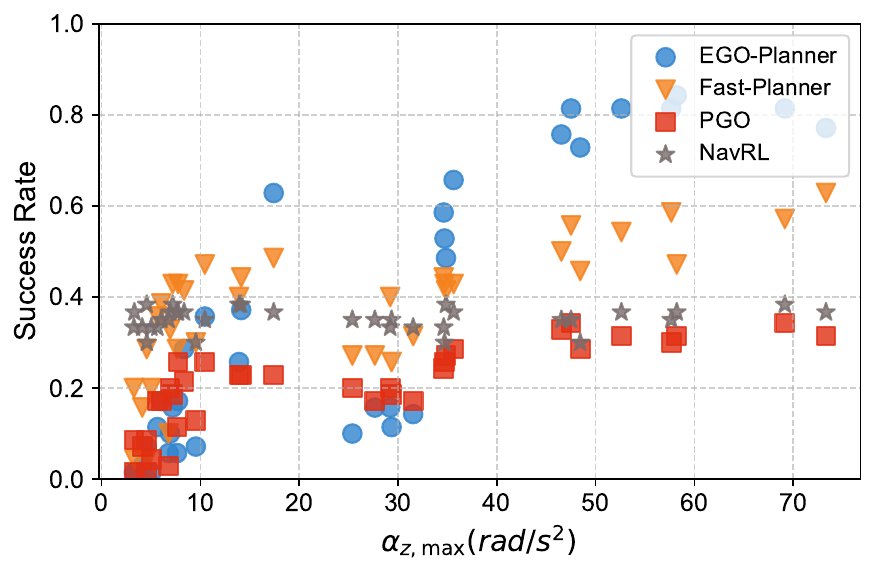}
\label{success_rate_analysis_3}}
\hfil
\caption{{\bf{Per-algorithm success rate vs. kinodynamic capability.}} Panels (a)-(c) plot each algorithm's success rate against the three kinodynamic indicators: (a) $\mathrm{TWR}_{\max}$, (b) $\alpha_{xy,\mathrm{max}}$ (log scale), and (c) $\alpha_{z,\mathrm{max}}$. Each marker represents the mean success rate for a single platform (averaged across all scenarios), with marker color indicating the algorithm as shown in the legend.}
\label{success_rate_analysis}
\end{figure*}

\begin{figure*}[!t]
\centering
\includegraphics[width=6.5in]{"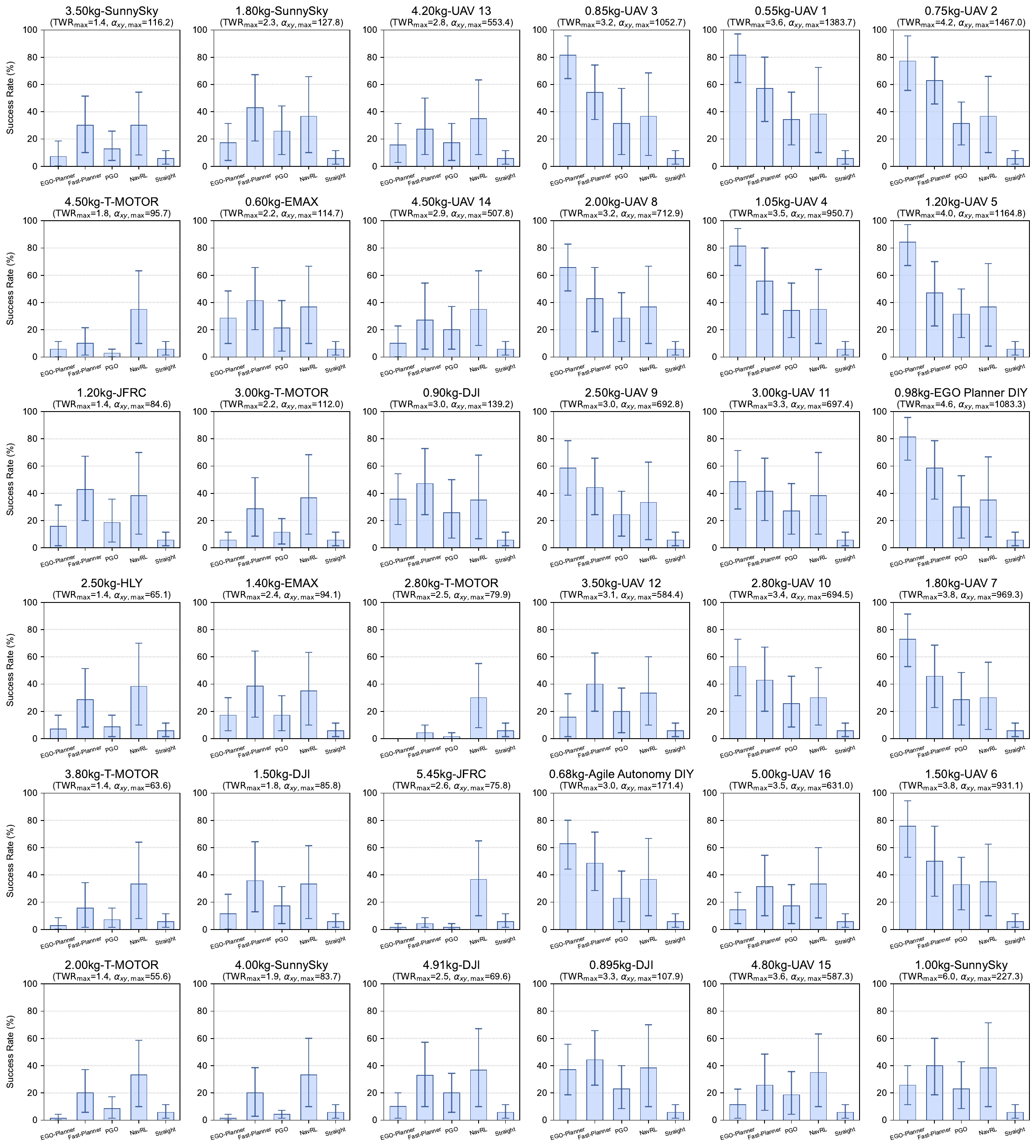"}
\caption{{\bf{Benchmark results by UAV kinodynamic profile.}} This figure shows the mean navigation success rate for each method across all 36 UAV platforms. The 36 subplots are strategically arranged in a 2D grid based on platform kinodynamics: horizontal position (left-to-right) corresponds to increasing $\mathrm{TWR}_{\max}$, and vertical position (bottom-to-top) corresponds to increasing $\alpha_{xy,\mathrm{max}}$. Each individual bar represents a method's average success rate across all seven test scenarios for that specific platform. Error bars indicate 95\% bootstrap confidence intervals.}
\label{different_dynamics}
\end{figure*}

\begin{figure}[!h]
\centering
\includegraphics[width=3.1in]{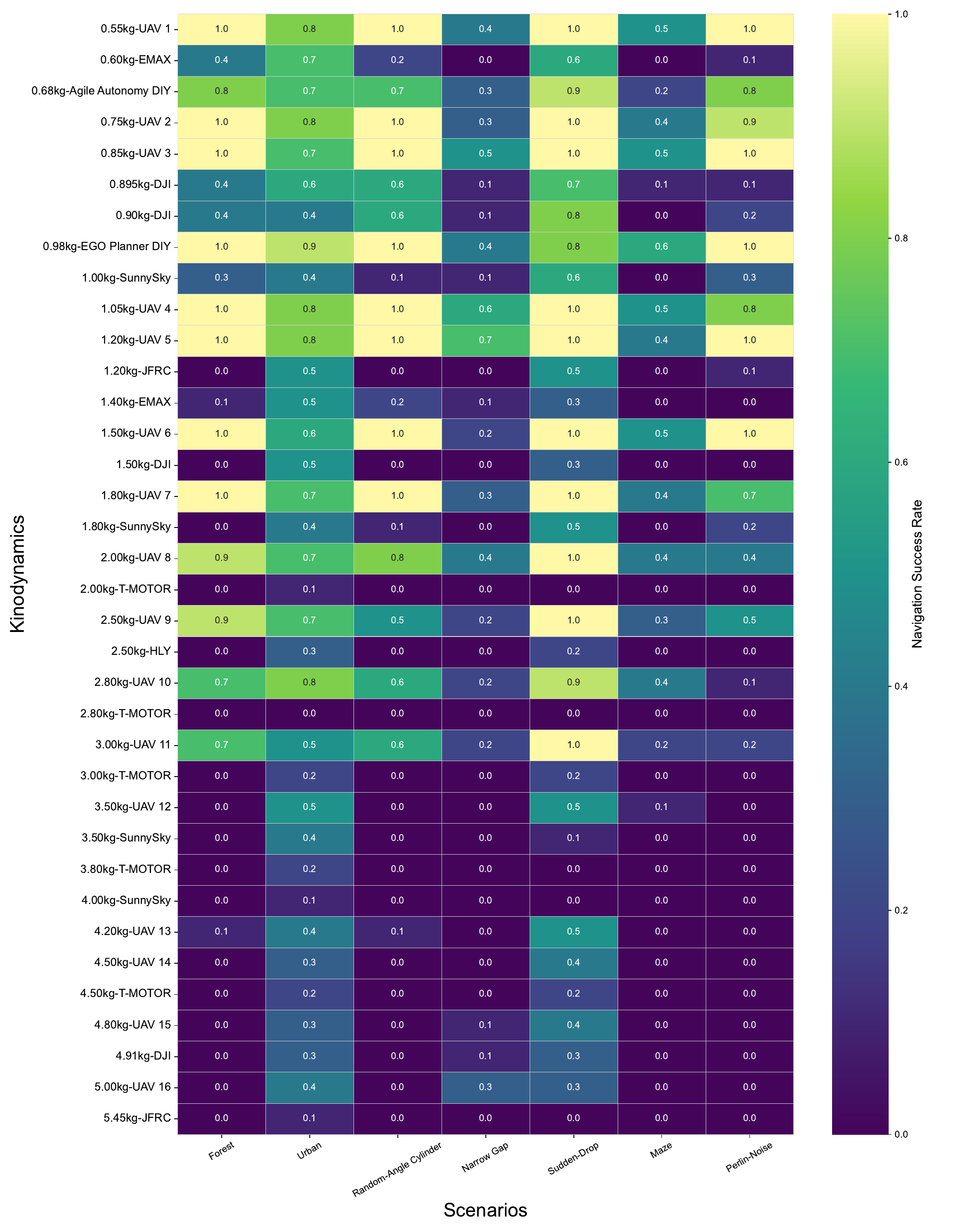}
\caption{{\bf{EGO-Planner navigation success rate.}} The heatmap shows the mean success rate for each platform (y-axis) across all seven scenarios (x-axis). Success rates are color-coded from 0.0 (dark purple) to 1.0 (bright yellow).}
\label{EGO_planner_Nav_success_rate}
\end{figure}

\begin{figure}[!h]
\centering
\includegraphics[width=3.1in]{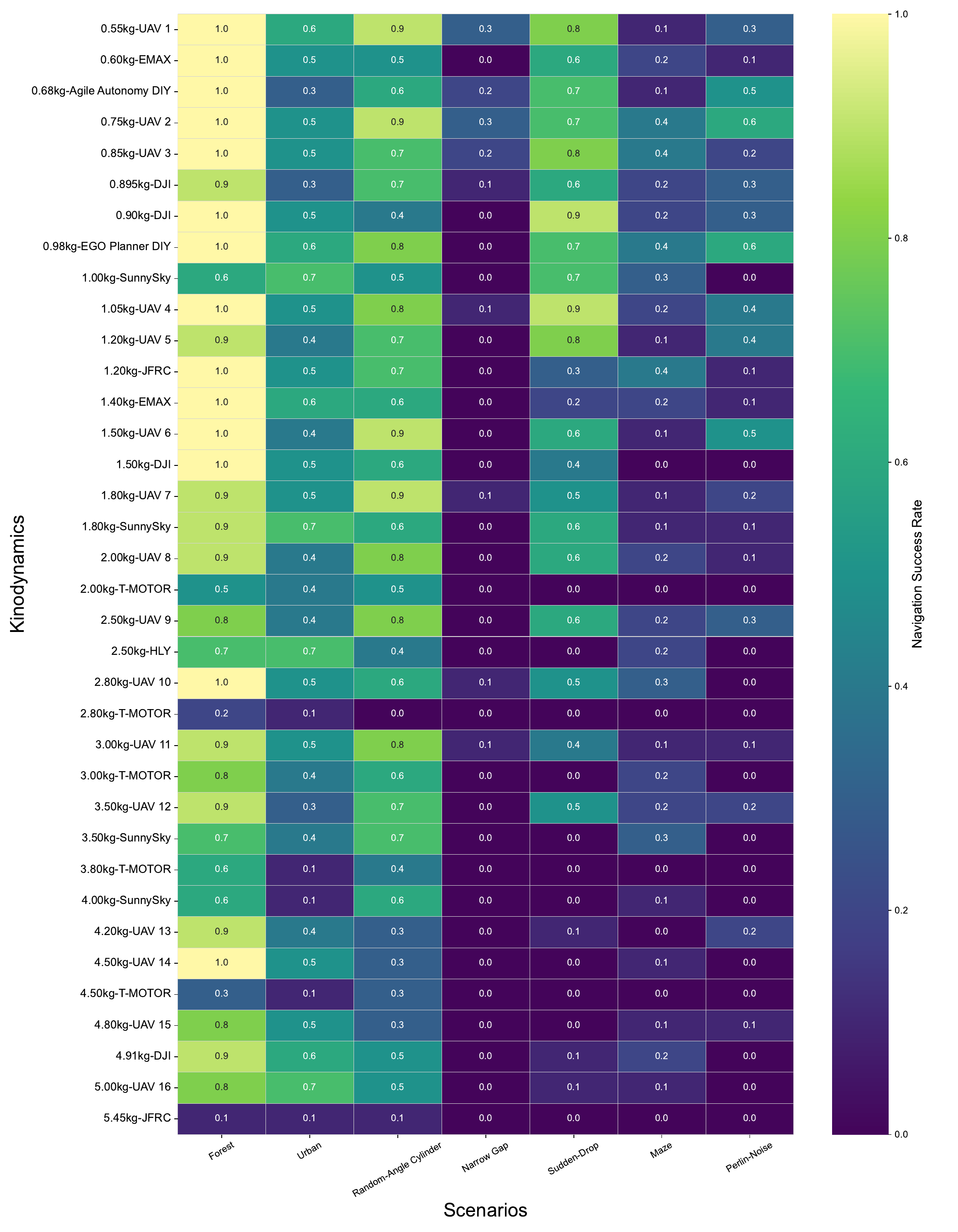}
\caption{{\bf{Fast-Planner navigation success rate.}} The heatmap shows the mean success rate for each platform (y-axis) across all seven scenarios (x-axis). Success rates are color-coded from 0.0 (dark purple) to 1.0 (bright yellow).}
\label{Fast_planner_Nav_success_rate}
\end{figure}

\begin{figure}[!h]
\centering
\includegraphics[width=3.1in]{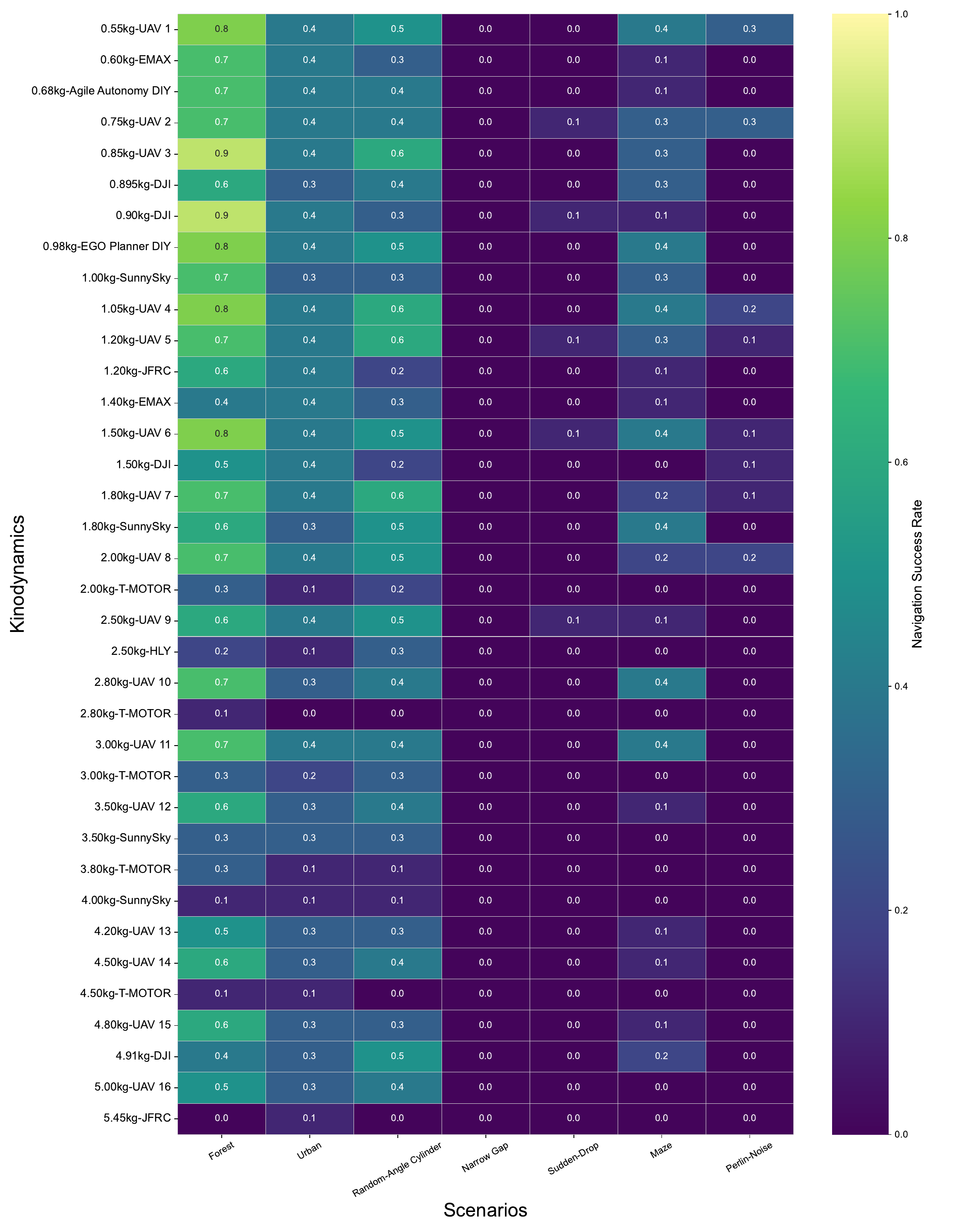}
\caption{{\bf{PGO navigation success rate.}} The heatmap shows the mean success rate for each platform (y-axis) across all seven scenarios (x-axis). Success rates are color-coded from 0.0 (dark purple) to 1.0 (bright yellow).}
\label{PGO_Nav_success_rate}
\end{figure}

\begin{figure}[!h]
\centering
\includegraphics[width=3.1in]{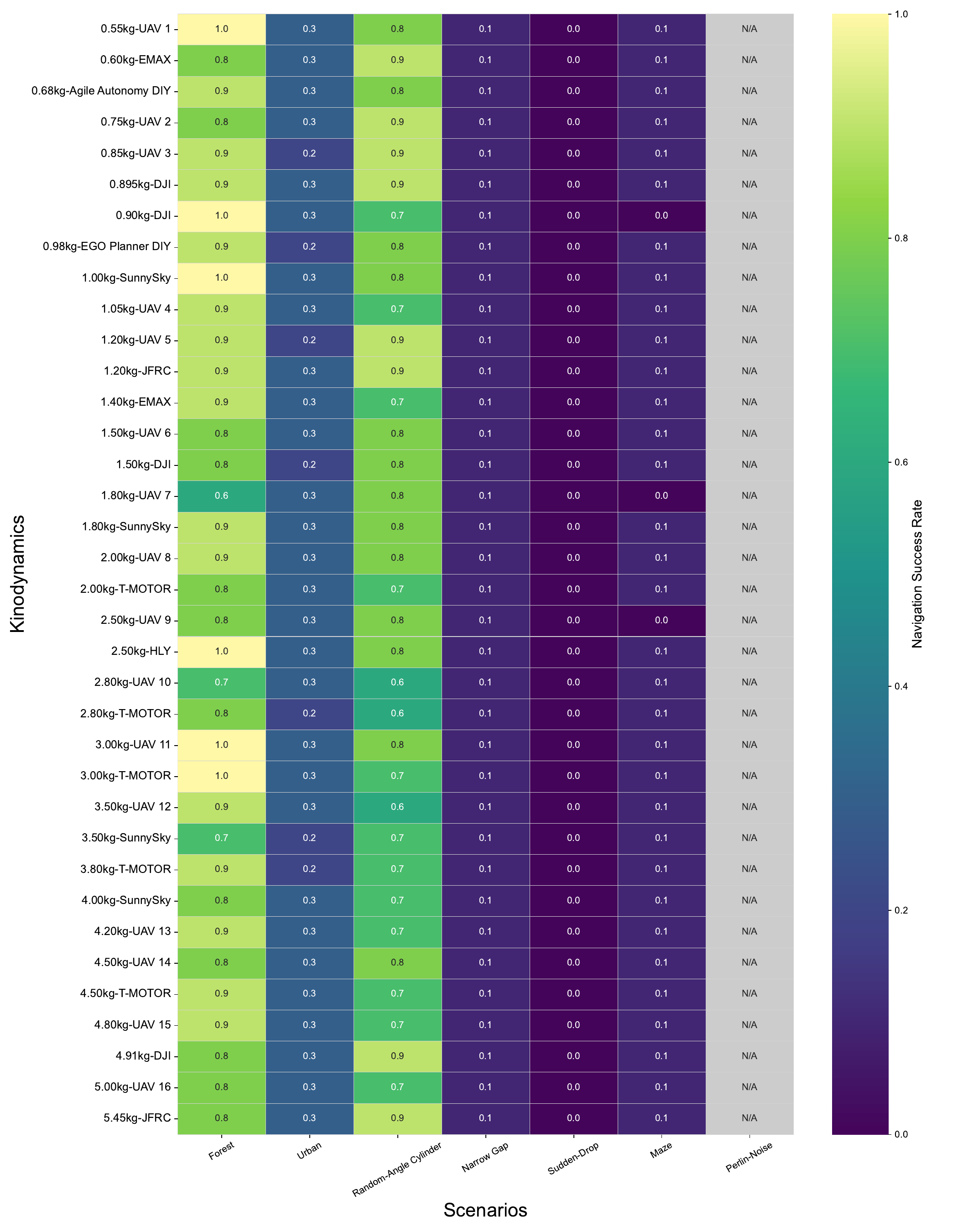}
\caption{{\bf{NavRL navigation success rate.}} The heatmap shows the mean success rate for each platform (y-axis) across all seven scenarios (x-axis). Success rates are color-coded from 0.0 (dark purple) to 1.0 (bright yellow).}
\label{NavRL_Nav_success_rate}
\end{figure}

\begin{figure}[!h]
\centering
\includegraphics[width=3.1in]{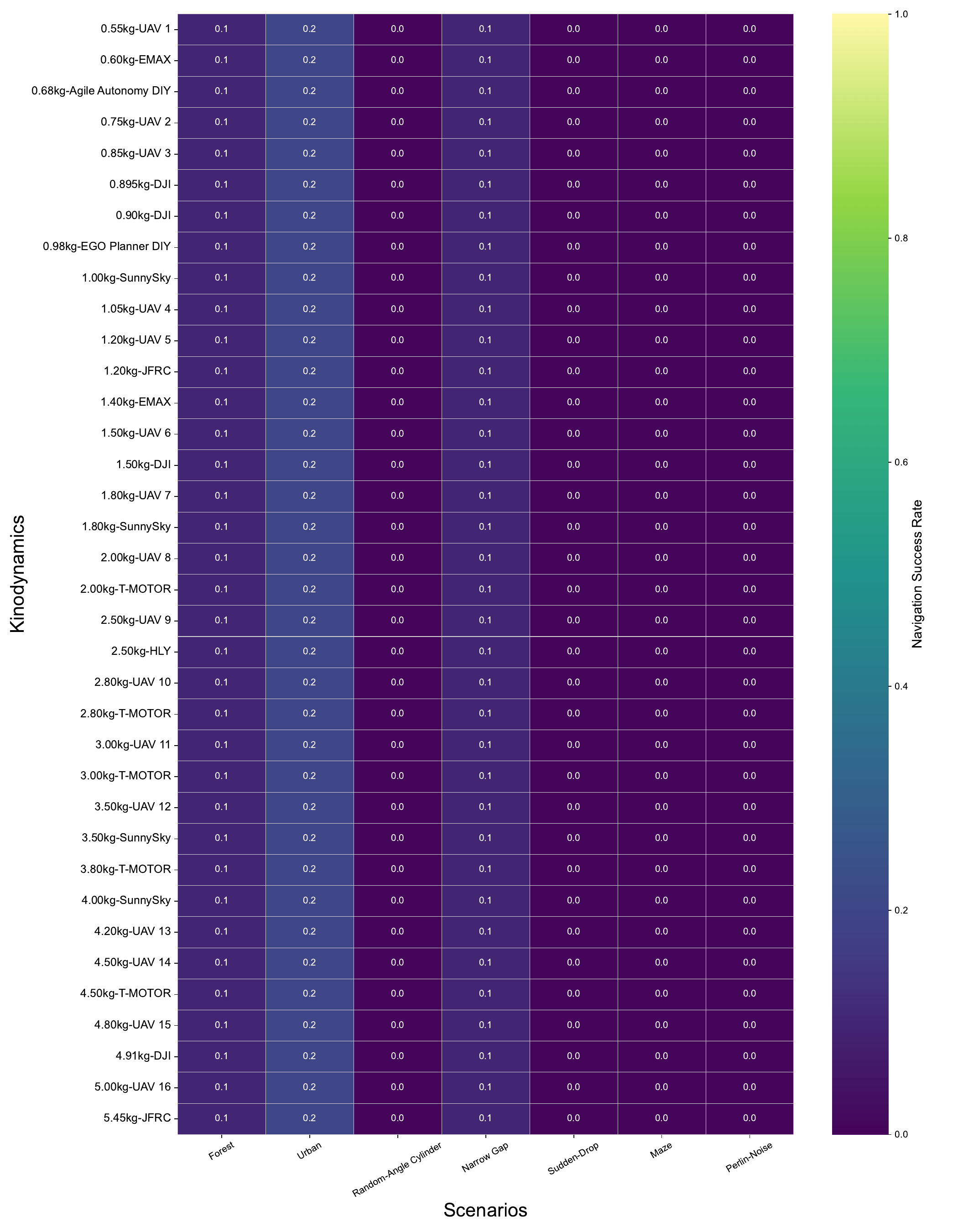}
\caption{{\bf{Straight-Flight Baseline navigation success rate.}} The heatmap shows the mean success rate for each platform (y-axis) across all seven scenarios (x-axis). Success rates are color-coded from 0.0 (dark purple) to 1.0 (bright yellow).}
\label{Straight_Nav_success_rate}
\end{figure}

\Cref{different_scenarios} summarizes per-scenario mean navigation success rates for all methods together with 95\% bootstrap confidence intervals computed from $B=1000$ resamples; each (algorithm, scenario, platform) combination is evaluated with 10 independent trials. Building on the scenario overview, \Cref{success_rate_analysis} reports algorithm performance aggregated across the heterogeneous set of UAV platforms to expose sensitivity to platform capability. \Cref{different_dynamics} complements this view by showing per-environment performance averaged across all platforms. For readers interested in fine-grained behavior, \Cref{EGO_planner_Nav_success_rate,Fast_planner_Nav_success_rate,PGO_Nav_success_rate,NavRL_Nav_success_rate,Straight_Nav_success_rate} provide per-algorithm heatmaps and additional diagnostics that reveal where and how each method succeeds or fails. Confidence intervals follow the bootstrap procedure of Efron and Tibshirani~\cite{tibshirani1993introduction}. We discuss the results in a Q\&A format.

{\bf{Q1: Which method performs best in each navigation scenario?}}

{\bf{A1:}} \Cref{different_scenarios} summarizes per-scenario mean success rates across all methods; \Cref{EGO_planner_Nav_success_rate,Fast_planner_Nav_success_rate,PGO_Nav_success_rate,NavRL_Nav_success_rate,Straight_Nav_success_rate} present per-algorithm heatmaps and additional diagnostics. Based on these plots, we compare methods scenario by scenario.

\begin{itemize}
  \item {{\bf{Open fields (Forest, Urban):}} In the relatively open Forest and Urban scenarios with sparse obstacles, all algorithms perform well. Fast-Planner and NavRL both exceed 80\% average success in the Forest scenario, with narrow confidence intervals indicating excellent stability and robustness. EGO-Planner and PGO lag slightly behind but still outperform the straight-flight baseline.}
  \item {{\bf{Random-Angle Cylinder:}} In the Random-Angle Cylinder scenario, obstacles are randomly oriented. NavRL leads with over 75\% average success and a tight confidence interval, Fast-Planner maintains above 55\%, while EGO-Planner and PGO drop to around 35\%. The straight-flight baseline fails entirely.} 
  \item {{\bf{Narrow-Gap:}} The Narrow-Gap scenario demands aggressive yaw maneuvers in confined spaces, resulting in generally low success rates. EGO-Planner slightly outperforms the baseline, NavRL matches the baseline, and Fast-Planner along with PGO perform worse than the baseline, indicating shortcomings in fine attitude control.}
  \item {{\bf{Sudden-Drop:}} The Sudden-Drop scenario tests vertical pitch maneuverability. EGO-Planner performs best at 55\% average success, followed by Fast-Planner at 38\%. PGO and NavRL both fall below 5\%, and the straight-flight baseline fails entirely, highlighting their lack of adaptation to pitch-intensive maneuvers.}
  \item {{\bf{Maze:}} The Maze scenario evaluates exploration and path-finding skills. EGO-Planner, Fast-Planner, and PGO each achieve around 15\% average success with similar confidence intervals, NavRL falls below 10\%, and the straight-flight baseline fails. This indicates that complex labyrinth tasks challenge all tested methods.}
  \item {{\bf{Random Perlin-Noise:}} In the most complex Random Perlin-Noise scenario, only classical planners are evaluated. EGO-Planner leads with 28\% success, Fast-Planner achieves 15\%, PGO drops to 4\%, and the straight-flight baseline fails entirely, illustrating differences in local mapping and curve optimization efficacy under extreme randomness.}
\end{itemize}

{\bf{Q2: How do UAV kinodynamic properties affect the success rates of visual navigation algorithms, and do the chosen indicators form a reasonable set of kinodynamic descriptors?}}

{\bf{A2:}} Conventional optimization-based methods (EGO-Planner, Fast-Planner) tend to achieve higher success rates on platforms with larger $\mathrm{TWR}_{\max}$ and greater $\alpha_{xy,\mathrm{max}}$ (see \Cref{success_rate_analysis_1} and \Cref{success_rate_analysis_2}). Trajectory optimizers produce aggressive, time-critical maneuvers that require both ample translational acceleration and fast reorientation of thrust (roll/pitch agility); platforms lacking these capabilities cannot reliably execute those trajectories. NavRL shows a comparatively flat response across these kinodynamic dimensions in our tests, suggesting either the learned policy cannot exploit higher agility or the training distribution did not emphasize high-agility regimes. $\alpha_{z,\mathrm{max}}$ plays a role in scenarios that require large heading changes, but its overall correlation with success is weaker than that of $\alpha_{xy,\mathrm{max}}$ (see \Cref{success_rate_analysis_3}).

Taken together, $\mathrm{TWR}_{\max}$, $\alpha_{xy,\mathrm{max}}$ and $\alpha_{z,\mathrm{max}}$ capture, respectively, translational power, thrust-direction reorientation capability and yaw responsiveness. As shown in \Cref{success_rate_analysis}, these three indicators form a compact and physically interpretable set that meaningfully constrains feasible navigation behaviors for quadrotors.

{\bf{Q3: Which method is most effective across different UAV platform?}}

{\bf{A3:}} \Cref{different_dynamics} summarizes per-platform average success rates grouped by platform category (virtual vs. real and by performance tier).

\begin{itemize}
  \item {{\bf{Virtual UAV Platforms:}} On high-performance virtual platforms, EGO-Planner achieves the highest average success rate, followed by Fast-Planner; PGO and NavRL perform similarly and both clearly outperform the straight-flight baseline. However, on low-performance virtual platforms, the average success rates of EGO-Planner, Fast-Planner, and PGO drop sharply and are no longer significantly better than the baseline. Only NavRL maintains a relatively stable success rate, indicating greater resilience to degraded kinodynamic capabilities.} 
  \item {{\bf{Real UAV Platforms:}} In real platform tests, NavRL delivers the highest and most consistent average success rates, remaining virtually unaffected by drops in platform performance. In contrast, EGO-Planner, Fast-Planner, and PGO not only achieve lower success rates overall but also suffer marked declines as platform performance decreases, performing comparably to the straight-flight baseline on low-end real platforms.} 
\end{itemize}

{\bf{Q4: How does the straight-flight baseline differ from active planners?}}

{\bf{A4:}} The straight-flight baseline flies at constant speed along the direct line from start to goal without collision avoidance. As shown in \Cref{different_scenarios}, it attains mean success rates of approximately 10-20\% in the open Forest and Urban scenarios, but fails in cluttered environments.

{\bf{Q5: Overall, which algorithm performs best?}}

{\bf{A5:}} Using the scoring method described in \Cref{scoring_method}, we computed a composite score for each visual UAV navigation algorithm. Weight choices were set as follows. Classic navigation scenarios were assigned an initial weight of 1.2, while theoretical scenarios were assigned 1.0. For platform weights, real UAV models were assigned 1.5 and virtual models 1.0. The stability penalty coefficient was set to $\beta = 0.3$.

\begin{table}[!ht]
\centering
\caption{Composite scores, variances and final scores for each algorithm.\label{tab2}}
\resizebox{\linewidth}{!}{
\begin{tabular}{l|c|c|c|c}
\toprule
\bf{Algorithm} & \bf{Score (no penalty)} & \bf{Variance} & \bf{FinalScore} & {\textbf{Missing scenarios}}\\
\midrule
EGO-Planner & 30.25 & 0.120 & 21.32 & - \\
Fast-Planner & 37.34 & 0.106 & $\mathbf{27.58}$ & - \\
Path-Guided PGO & 20.39 & 0.054 & 17.70 & - \\
NavRL & 37.78$^\dagger$ & 0.122 & 26.41$^\dagger$ & Perlin-Noise \\
Straight Flight Baseline & 6.05 & 0.005 & 5.97  & - \\
\bottomrule
\end{tabular}
}
\end{table}

The computed scores appear in \Cref{tab2}, As noted, $\dagger$NavRL has no valid results for the Perlin-Noise scenario. The NavRL score shown here is computed after excluding that scenario and renormalizing the remaining scenario weights (see \Cref{scoring_method}). Because the effective evaluation set differs, NavRL’s score is provided as a reference only and is not directly comparable to scores computed over the full scenario set.

Overall, Fast-Planner attains the highest composite score, balancing high success rates with relatively low variance across scenarios and platforms. The ranking from highest to lowest score is: Fast-Planner, EGO-Planner, Path-Guided PGO, Straight Flight Baseline. NavRL’s reference score is also relatively high, but its performance exhibits substantially larger variance, indicating weaker stability compared with the other methods.

{\bf{Q6: What are the strengths and weaknesses of optimization-based and learning-based navigation methods?}}

{\bf{A6:}} From the benchmarking of representative algorithms, we can distill the respective strengths and weaknesses of optimization-based and learning-based navigation methods:

{\bf{Optimization-based methods}} (e.g., Fast-Planner, EGO-Planner, PGO) explicitly encode vehicle dynamics, safety, and efficiency constraints in trajectory planning, yielding several advantages:

\begin{itemize}
  \item {{\bf{High performance utilization:}} These methods consistently reach the $4~\mathrm{m/s}$ speed limit and generate dynamically feasible trajectories involving full 3D maneuvers, such as pitch and yaw. This enables superior performance in demanding scenarios like Sudden-Drop and Narrow-Gap.} 
  \item {{\bf{Physically interpretable and generalizable:}} The model-based design allows predictable, tunable behavior, facilitating deployment in real-world applications.} 
  \item {{\bf{Strong controllability and stability:}} Their reliance on analytical solvers makes them robust to runtime fluctuations in most structured scenarios.}
\end{itemize}

However, limitations include:

\begin{itemize}
  \item {{\bf{Dependence on accurate models and sensors:}} Any model mismatch or sensor noise can lead to suboptimal paths or even navigation failure.} 
  \item {{\bf{Vulnerability to local minima:}} Especially in random or unstructured scenarios, their gradient-based optimization may struggle to escape poor initial paths.} 
\end{itemize}

{\bf{Learning-based methods}}, as represented by NavRL in our study, offer different trade-offs. While these findings highlight potential patterns in data-driven approaches, it is crucial to recognize that they are based on a single pre-trained model and may not generalize to all learning-based systems.

\begin{itemize}
  \item {{\bf{Platform-agnostic performance:}} NavRL maintains stable success rates even on low-performance drones, highlighting its resilience across kinodynamic variations.} 
  \item {{\bf{Strong sim-to-real transfer:}} Without retraining or parameter tuning, NavRL is able to navigate successfully in the Gazebo simulator using the pre-trained policy, highlighting its practical transfer potential and limited generalization to familiar settings.} 
\end{itemize}

However, we also identified key drawbacks that may stem from its specific implementation and training:

\begin{itemize}
  \item {{\bf{Underutilization of flight capability:}} NavRL consistently capped at $1~\mathrm{m/s}$ despite the $4~\mathrm{m/s}$ limit, indicating poor dynamic exploitation.} 
  \item {{\bf{Behavioral bias from training data:}} The policy exhibits only right-turn obstacle avoidance and lacks pitch maneuvers, likely due to skewed training data lacking diverse motion examples. This causes complete failure in tests like Narrow-Gap and Sudden-Drop.}
\end{itemize}

\section{Broader Impact and Limitations}
\label{Broader Impact and Limitations}
Although FLYINGTRUST covers a broad spectrum of kinodynamic profiles and navigation scenarios, it currently fixes control-loop parameters (e.g., PID/MPC gains) and does not measure how low-level controller tuning affects higher-level planning performance. Future extensions should:

\begin{itemize}
  \item {{\bf{Incorporate control-parameter sensitivity:}} Systematically vary controller gains and estimators to quantify their impact on success rate, trajectory smoothness, and replanning latency.} 
  \item {{\bf{Model actuator and sensor uncertainties:}} Integrate noise models, time delays, and hardware degradation into simulation to better approximate real-world conditions.}
  \item {{\bf{Expand scenario diversity:}} Add dynamic obstacles and lighting variations to stress-test perception and planning modules.}
\end{itemize}

Our comparative analysis uncovers clear trade-offs between model-based optimization and data-driven learning. To advance quadrotor visual navigation, we recommend three complementary directions.

\begin{itemize}
  \item {{\bf{Hybrid planning and learning:}} Combine real-time optimization or topological guidance with learned policy priors so that principled constraints provide safety and feasibility while data-driven components supply adaptability and local refinement. For example, use an optimizer to generate a safe guide path and a learned policy to track or refine that guide under perception noise.} 
  \item {{\bf{Kinodynamic integration:}} Embed platform-specific thrust and inertia models into cost functions or learning objectives so that generated trajectories and learned behaviors respect physical limits and remain implementable on target hardware.}
  \item {{\bf{Focus on generalization:}} Train and validate on procedurally generated, diverse obstacle layouts and heterogeneous vehicle profiles to reduce overfitting to narrow distributions and to improve cross-platform and cross-scene robustness.}
\end{itemize}

\section*{Acknowledgments}
This work was supported by the National Natural Science Foundation of China under Grant No.T2350005 and the Fundamental Research Funds for the Central Universities, Sun Yat-sen University under Grant No.23xkjc008.

\bibliographystyle{IEEEtran}
\bibliography{ref_v1}

\newpage

\end{document}